\newenvironment{customlegend}[1][]{%
    \begingroup
    \csname pgfplots@init@cleared@structures\endcsname
    \pgfplotsset{#1}%
}{%
    \csname pgfplots@createlegend\endcsname
    \endgroup
}%

\def\addlegendimage{\csname pgfplots@addlegendimage\endcsname}

\documentclass[journal]{IEEEtran}
\ifCLASSINFOpdf
\else
\fi
\usepackage{mathrsfs}
\usepackage{mathtools,amsfonts,amssymb,amsthm}
\usepackage{footnote}
\makesavenoteenv{tabular}
\makesavenoteenv{table}
\usepackage{mathtools}
\usepackage{amsmath}
\usepackage[english]{babel}
\usepackage{relsize}
\usepackage{lipsum}
\usepackage{algorithm}
\usepackage{algorithmicx}
\usepackage{algpseudocode}
\usepackage{lipsum,graphicx,multicol}
\usepackage{float}
\usepackage{setspace}
\usepackage{tablefootnote}
\usepackage[T1]{fontenc}
\usepackage{color, colortbl}
\usepackage{booktabs}
\usepackage{tikz}
\usetikzlibrary{fit,matrix}
\usetikzlibrary{calc}
\usepackage{bm}
\usepackage{amssymb}
\usepackage{makecell}
\usepackage{pgfplots}
\usepackage{threeparttable}
\usepackage{caption}
\usepackage{subcaption}
\usepackage[skins]{tcolorbox}
\usepackage{mathrsfs}
\usepackage{tabularx}
\usepackage{array}
\usepackage{cite}
\newcolumntype{C}{>{\centering\arraybackslash}p{28.5em}}

\usepackage[utf8]{inputenc}
\usepackage{fourier} 
\usepackage{array}
\usepackage{makecell}

\DeclareMathOperator*{\argmax}{argmax} 
\DeclareMathAlphabet{\mathcal}{OMS}{cmsy}{m}{n}

\newtheorem{proposition}{Proposition}
\theoremstyle{definition} \newtheorem{definition}{Definition}
\newcommand{\eapcomment}[1]{\textcolor{black}{ #1}}

\newcommand{\iapcomment}[1]{\textcolor{black}{ #1}}

\begin{document}
%
\title{Tractable Learning and Inference for Large-Scale Probabilistic Boolean Networks\\}
%
%
%

\author{Ifigeneia Apostolopoulou,
        Diana Marculescu,~\IEEEmembership{Fellow,~IEEE}
\thanks{Ms. Apostolopoulou is with the Machine Learning Department, Carnegie Mellon University, Pittsburgh,
PA, 15213 (e-mail: iapostol@andrew.cmu.edu).}
\thanks{Ms. Marculescu is with the department of Electrical and Computer Engineering, Carnegie Mellon University, Pittsburgh,
PA, 15213 (e-mail: dianam@andrew.cmu.edu).}
}

\maketitle

%
%
%
\begin{abstract}
Probabilistic Boolean Networks (PBNs) have been \eapcomment{previously proposed} so as to gain insights into complex dynamical systems. However, identification of large networks and of the underlying discrete Markov Chain which describes \eapcomment{their} temporal evolution, still remains a challenge. In this paper, we introduce an equivalent representation \eapcomment{for the PBN}, the Stochastic Conjunctive Normal Form \eapcomment{(SCNF)}, \eapcomment{which} paves the way to a scalable learning algorithm and helps predict long-run dynamic behavior of large-scale systems. Moreover, \eapcomment{SCNF} allows \iapcomment{its} efficient sampling \eapcomment{so as to} statistically infer multi-step transition probabilities \iapcomment{which can provide knowledge on the activity levels of individual nodes in the long run}.
\end{abstract}

%
\IEEEpeerreviewmaketitle

\section{Introduction}
%
%
%
%


\IEEEPARstart{M}{any} complex systems that have recently received intense research attention by the scientific community fall in the broad category of \textit{dynamical systems}\cite{liang2012robust,yang2017exponential,zhang2011data, li2016observer, li2016observerb, xu2014adaptive}. A dynamical system is typically governed by rules that describe the time dependence of a set of variables. 
\textit{Nonlinear} (\textit{Linear}) dynamical systems which are, in general, described by a system of nonlinear (linear) differential equations, can often be reconstructed and analyzed using a qualitative or semi-qualitative estimate of the behavior of its state variables. In situations where the exact values of the states are not required, estimating the behavior of such systems using binary logical models can be extremely fast when compared to learning and simulating complicated systems of differential equations using numerical methods.

Boolean Networks (BNs), originally proposed by Kauffman \cite{kauffman1969metabolic,stuart1993origins}, constitute a very-well studied qualitative modeling framework, and has been used for multifarious applications such as genetic regulatory networks \cite{kauffman1969metabolic, heidel2003finding}, neural networks \cite{wang1990oscillations, kurten1988correspondence}, BN robots \cite{roli2011design}, econometrics \cite{caetano2015boolean} among others. The Probabilistic Boolean Network (PBN) paradigm was introduced by Shmulevich  \cite{shmulevich2002probabilistic} as a semi-qualitative extension of BN for an alternative representation of gene regulatory networks \cite{shmulevich2002boolean}; it combines rule-based modeling with uncertainty principles. PBNs have been deployed in a spectrum of applications similar to those of BNs \cite{trairatphisan2013recent,liang2014construction}.

For the past decade, BNs and PBNs have been the object of extensive studies. Past theoretical studies focus on learning \cite{marshall2007inference,ching2005construction,ching2009generating}, steady-state distribution analysis \cite{ching2007approximation}, relations to Dynamic Bayesian networks \cite{murphy2002dynamic}, \cite{lahdesmaki2006relationships}, and Markov Chains \cite{anderson1957statistical}, \cite{shmulevich2002probabilistic}, \cite{xiao2009tutorial}. Recent research efforts have been primarily focused on control \cite{wu2017policy,zhao2016control,liu2017delayed,liu2017normalization,li2017single, li2016pinning, zhang2013controllability, layek2009adaptive, qian2010state}, synchronization \cite{chen2017synchronization,li2013synchronization, zhang2016synchronization,zhong2014synchronization}, steady state analysis \cite{mizera2016parallel},\cite{assa} and network identification \cite{melkman2017identifying,cheng2016exact,zhang2017identification,liu2008inference,kobayashi2017design}. 
\iapcomment{Identification of the long-run behavior of dynamical systems is of utter importance since it usually conveys domain implications. For instance, the characterization of the network's long-run dynamics plays a crucial role in treatment of various human cancers such as breast cancer, and
leukemia \cite{choi2012attractor, saadatpour2011dynamical, hupp2000strategies}. Furthermore, it is possible to control certain nodes in a
network, such that the whole system can evolve according to a desirable probability distribution \cite{ostergaard2000increasing}. However, automated }system reconstruction faces main challenges. These hindrances arise mostly from the exponential growth of possible model configurations and the limited observations under changing initial conditions. 

Our contributions revolve around \textit{model learning} and \textit{dynamics inference} for PBNs under an alternative representation. This formulation leads to an Approximate Maximum Likelihood Estimation (AMLE) method which renders the tackling of both problems at a large scale feasible. To the best of our knowledge, no prior research has managed to address successfully the reconstruction of the state evolution of general PBNs in the order of 1000 nodes (or equivalently a Markov Chain with $2^{1000}$ states), as opposed to other exact likelihood \cite{dumcke2013exact} and information-theoretic approaches \cite{berestovsky2013evaluation} which can handle only deterministic and small-scale dynamics. \iapcomment{Thus, our method provides a new framework for prediction of temporal dynamics generated from large networks, a problem crucial to modeling gene regulation, cell signaling, and other complex mechanisms.}
%
%

\section {Related Work}
As already been pointed out in prior work, the dynamical behavior of a PBN can be described by Markov Chain theory and thus, tools developed for that can be applied to the analysis of PBNs. A Maximum Likelihood Estimation (MLE) approach for estimating the transition probability matrix of the Markov Chain (and the associated PBN) is presented in \cite{anderson1957statistical}, along with \eapcomment{certain} theoretical guarantees. However, as expected, straightforward transition probabilistic representations require the estimation of $2^N\times(2^N-1)$ probabilities, where $N$ is the number of nodes in the network. Therefore, \eapcomment{such representations} demand an unrealistic amount of data which hinders their adoption in real-world scenarios. Other studies \cite{ching2005construction} also use Markov chains for solving the problem of predicting the system dynamics. However, the state probability of individual nodes is represented as a linear combination of $N$ $2 \times 2$ (in the case of binary logic) transition probability matrices, 
which pertain to the influence of only one node on the dynamics of the targeted node. Hence, while the number of parameters that have to be estimated is reduced to $O(N^2)$, the problem can still be considered quite complex for large $N$. Moreover, the equivalent PBN can be learned as a $2^N \times 2^N$ transition probability matrix, from which it is hard to extract the underlying logic rules. This size can be prohibitively large for computing multi-step transition probabilities and for steady-state analysis which can provide valuable information for developing intervention-oriented approaches \cite{shmulevich2002gene, shmulevich2003steady}. 

Other work \cite{ching2009generating} expresses the transition probability matrix as a sum of Boolean Network matrices and estimates the selection probabilities of the rules, while assuming that the transition probability matrix and the logic portion of the PBN are known. The learning procedure in \cite{marshall2007inference} attempts to learn both the logic portion (in the form of a truth table) and the parametric portion (in the form of switching, selection, and perturbation probabilities). However, the amount of temporal data needed for estimating the parametric part, which is crucial for dynamics recovery, is huge. Indeed, only results for network connectivity are reported, and only for network sizes of up to 7 variables. 
Moreover, a maximum, much \eapcomment{smaller} than $N$ number of different interacting nodes/variables (\textit{node in-degree}) in the Boolean functions known \textit{a priori} is assumed,  while, as already mentioned, the tabular representation of the learned PBN is unsuitable for further analysis. The approach described in \cite{pahuja2016learning} relies on prior domain knowledge in terms of the biological pathways of the network that has to be learned as a PBN, and it can cope with only up to 7 nodes. Similarly, the software tools described in \cite{trairatphisan2014optpbn}, \cite{mizera2017assa} require prior knowledge on the possible interactions between the nodes. The work in \cite{cheng2016exact} and \cite{melkman2017identifying} offers sample complexity guarantees for PBNs and threshold PBNs respectively, both of which only pertain to the  discovery of the logic part. Moreover, conditions on the number of the constituent boolean functions (which come only in the form of pairs or triplets), the logic structure of the rules (only AND/OR boolean functions), and fixed node in-degrees are necessary for the derivation of these results.


The main contributions of this article include: 
\begin{enumerate}
\item We propose the SCNF (Stochastic Conjunctive Normal Form) network \eapcomment{as} an alternative representation for PBNs.
\item We suggest a scalable  and accurate learning algorithm, which manages to recover both the logic and the parametric portion of the underlying PBN from a sufficiently small number of observed system transitions, without making any prior assumptions on the structure of the logic formulas that have to be learned, and without using any prior domain knowledge.
\item  We show that the SCNF model is amenable to efficient stochastic simulation, and can therefore be used to infer approximately up to \eapcomment{100-}step system transition probabilities.
\end{enumerate}
The formal definitions of BNs and PBNs are given in Section III. In Section IV, we present our model definition and show its equivalence to PBNs. In Section V, we provide our reconstruction algorithm. 
In Section VI, we present experimental results which demonstrate accurate dynamic prediction of new temporal trajectories and efficient transition probabilities estimation. Finally, Section VII provides the reader with examples which illustrate the SCNF definition and learning.
%
%
\section{Preliminaries}

\subsection{Boolean Networks (BNs)}
\begin{definition}
\setstretch{1.35}
A Boolean Network (BN) is a directed network with $N$ binary-valued nodes $V=\left\{x_1, x_2, \dots, x_N\right\}$. Each node $x_i$ has $N^{(i)}$ parent nodes $V^{(i)}=\big\{x^{(i)}_1, x^{(i)}_2, \dots, x^{(i)}_{N^{(i)}}\big\}$, where $x^{(i)}_j \in V$.  Let $s_t(x) \in \left\{False, True\right\}$ be the state of node $x$ at time $t$. Define $s^{(i)}_t \triangleq s_t(x_i)$. The state of node $x_i$ is regulated by a boolean function $f^{(i)}: \{False, True \}^{N^{(i)}} \rightarrow \{False, True\}$, such that $f^{(i)}=f^{(i)}\big(x^{(i)}_1, x^{(i)}_2, \dots, x^{(i)}_{N^{(i)}}\big)$. The state of the whole network at time step $t$ is represented by the vector $\mathbf{S}_t=\big[s_t^{(1)},s_t^{(2)},\dots, s_t^{(N)}\big] \in \{False, True\}^N$. Let $\mathbf{F}=\big[f^{(1)}, f^{(2)}, \dots, f^{(N)}\big]$ be the network function. The states of all nodes are updated synchronously (at the same time). Then the dynamics of the BN is given by $\mathbf{S}_{t+1}=\mathbf{F}(\mathbf{S}_t)$, where $s_{t+1}^{(i)}=f^{(i)}\big(\mathbf{s}_t\big(x^{(i)}_1\big),\dots,\mathbf{s}_t\big(x^{(i)}_{N^{(i)}}\big)\big)$ for $i=1,2,\dots,N$. We denote the BN by $G(V, \mathbf{F})$.
\end{definition}
Note that the network function is homogeneous in time, meaning that it is time invariant. Therefore, we can drop the time quantifier and the dynamics equation can be further simplified to $\mathbf{S}'=\mathbf{F}(\mathbf{S})$ (with $\mathbf{S}'$ representing the next state of the system and $\mathbf{S}$ the current state). The initial state (or initial condition) $\mathbf{S}_0$ and the network function $\mathbf{F}$ fully determine the evolution of the BN: $\mathbf{S}_0 \rightarrow \mathbf{S}_1 \rightarrow \dots \rightarrow \mathbf{S}_t \rightarrow \dots \rightarrow$. 
\subsection{Probabilistic Boolean Networks (PBNs)}
A Probabilistic Boolean Network (PBN) is an extension of a BN:
\begin{definition}
\setstretch{1.35}
A Probabilistic Boolean Network (PBN) is a directed network with $N$ binary-valued nodes $V=\{x_1, x_2, \dots, x_N\}$.
Each node $x_i$ has $N^{(i)}$ parent nodes $V^{(i)}=\big\{x^{(i)}_1, x^{(i)}_2, \dots, x^{(i)}_{N^{(i)}}\big\}$. Let $s_t(x) \in \big\{False, True\big\}$ be the state of node $x$ at time $t$. Define $s^{(i)}_t \triangleq s_t(x_i)$. The state of node $x_i$ is regulated by one Boolean function which is randomly selected from a set of $M^{(i)}$ Boolean functions $\mathbf{f}^{(i)}=[f^{(i)}_1, f^{(i)}_2,\dots, f^{(i)}_{M^{(i)}}]$ according to a categorical distribution $Categorical\big(M^{(i)},\mathbf{p}^{(i)}\big)$, where $\mathbf{p}^{(i)}=\big[p^{(i)}_1, p^{(i)}_2, \dots, p^{(i)}_{M^{(i)}}\big] \in [0,1]^{M^{(i)}}$, such that $\sum_{j=1}^{M^{(i)}}p^{(i)}_j=1$, and $p^{(i)}_j$ is the probability that the function $f^{(i)}_j$ will be selected. Each $f^{(i)}_{j}$ has $N^{(i)}_j$ variables, $f^{(i)}_j: \{False,True\}^{N^{(i)}_j} \rightarrow \{False, True\}$ such that it satisfies: $f^{(i)}_j=f^{(i)}_j\big(x^{(i)}_{j,1}, x^{(i)}_{j,2},\dots,x^{(i)}_{j,N^{(i)}_j}\big)$. If $V^{(i)}_j$ is the set of variables of the rule $f^{(i)}_j$, i.e, $V^{(i)}_j=\big\{x^{(i)}_{j,1}, x^{(i)}_{j,2},\dots,x^{(i)}_{j,N^{(i)}_j}\big\}$, it holds that the set of the $N^{(i)}$ parents of node $x_i$ will be $V^{(i)}=\bigcup_{j=1}^{M^{(i)}}V^{(i)}_j$, with $V^{(i)} \subseteq V$. The state of the whole network at time step $t$ is represented by the vector $\mathbf{S}_t=\big[s_t^{(1)},s_t^{(2)},\dots, s_t^{(N)}\big] \in \{False, True\}^N$. Let $\mathbf{F}=\big[\mathbf{f}^{(1)}, \mathbf{f}^{(2)}, \dots, \mathbf{f}^{(N)}\big]$ be the network function, $\mathbf{P}=\big[\mathbf{p}^{(1)}, \dots,\mathbf{p}^{(N)}\big]$ and $\tilde{\mathbf{F}}=(\mathbf{F},\mathbf{P})$. The states of all nodes are updated synchronously (at the same time) and independently. Then the dynamics of the PBN is given by $\mathbf{S}_{t+1}=\tilde{\mathbf{F}}(\mathbf{S}_t)$, where $s_{t+1}^{(i)}=f^{(i)}_j\big(\mathbf{s}_t\big(x^{(i)}_{j,1}\big),\dots,\mathbf{s}_t\big(x^{(i)}_{j,{N^{(i)}_j}}\big)\big)$ and $j \sim Categorical\left(M^{(i)},\mathbf{p}^{(i)}\right)$ for $i=1,2,\dots,N$. We denote the PBN by $G(V,\mathbf{\tilde{F}})$.
\end{definition}
Note that the categorical distributions are mutually independent, time invariant and independent of the past. Given the above definition, there are $M=\prod_{i=1}^{N}M^{(i)}$ constituent networks. The $j$-th network is described by $\mathbf{F}^j=\big[f^{(1)}_{j(1)}, f^{(2)}_{j(2)}, \dots, f^{(N)}_{j(N)}\big]$, with ${j(i)} \in \{1,2,\dots,M^{(i)}\}$  \iapcomment{denoting the boolean function selected for node $i$ in the Boolean Network $j$, and} is selected with probability $p(\mathbf{F}^j)=\prod_{i=1}^{N}p^{(i)}_{j(i)}$. Since the selections of boolean rules at time $t$ occur simultaneously, independently of  the other nodes and of the states in the past, $\mathbf{S}_{t-1},\;\mathbf{S}_{t-2},\;\dots$, a PBN generates a \iapcomment{discrete-time, }homogeneous, $2^N$-state Markov chain, which can be fully characterized by a transition probability matrix 
$\pmb{\mathscr{P}} \in \mathbb{R}^{{2^N}\times {2^N}}$, where the entry 
$\pmb{\mathscr{P}}(\mu, \lambda)$, with $\lambda,\mu \in \{0,1,\dots, 2^N-1\}$ represents the probability of moving  from state $\boldsymbol{\mu}$ to state $\boldsymbol{\lambda}$ , by considering $\boldsymbol{\lambda}, \boldsymbol{\mu} \in \{False, True\}^N$ as the boolean representation of the integers $\lambda, \mu$. The matrix $\pmb{\mathscr{P}}$ can be factorized as $\mathscr{P}=\sum_{j=1}^{M}p(\mathbf{F}^j)\pmb{\mathscr{A}}^j$, where $\pmb{\mathscr{A}}^j \in \mathbb{R}^{2^N \times 2^N}$ is the deterministic transition matrix of the Boolean Network $\mathbf{F}^j$. This decomposition yields $O(MN2^{2N})$ complexity for the computation of $\pmb{\mathscr{P}}$ \cite{liang2012stochastic}.

A graphical representation of $\pmb{\mathscr{P}}$ for a PBN is the state transition diagram. The reader may refer to Figure \ref{fig:example1} for a numerical example of a state transition diagram while the formal definition is given below:
\begin{definition}
\setstretch{1.35}
The state transition diagram of an $N$-node PBN $G(V,\mathbf{\tilde{F}})$ is a weighted directed graph 
$D(\mathcal{V}_s,\mathcal{E})$. $\mathcal{V}_s=\{False, True\}^N$ is a set of $2^N$ vertices, each representing a possible state of the PBN; $\mathscr{E}$ is a set of $2^{2N}$ edges, each pointing from a state $\boldsymbol{\mu} \in \mathcal{V}_s$ to its successor state $\boldsymbol{\lambda} \in \mathcal{V}_s$. Its weight is the probability of moving from $\boldsymbol{\mu}$ to $\boldsymbol{\lambda}$, as dictated by $\mathbf{F}$ and $\mathbf{P}$ of the PBN.
\end{definition}
\subsection{Problem Statement}
Assume that we observe $\mathsf{D}$, a list of $R$ boolean time series of potentially varying lengths $n_i$, for $i=1,2,\dots,R$: 
\begin{equation}
\mathsf{D}=\left(\big(\mathbf{S}^{1}_{t^1_0},\mathbf{S}^{1}_{t^1_1},\dots,\mathbf{S}^{1}_{t^1_{n_1}}\big),\dots, \big(\mathbf{S}^{R}_{t^R_0},\mathbf{S}^{R}_{t^R_1},\dots,\mathbf{S}^{R}_{t^R_{n_R}}\big)\right),
\end{equation}
where $\mathbf{S}^{r}_{t^r_k} \in \{False,True\}^N$. $\mathsf{D}$ can be transformed to a list (with repetitions of elements) of ordered pairs $(\mathbf{S},\mathbf{S}') \in \mathsf{L}$ which represent the transitions from the previous state $\mathbf{S}$ to the next state $\mathbf{S}'$ of the dynamical system:
\begin{equation}
\mathsf{L}\triangleq\mathlarger{\bigcup}\limits_{r=1}^{R}\left(\big(\mathbf{S}^r_{t^r_k}, \mathbf{S}^r_{t^r_{k+1}}\big)\right)_{k=0}^{n_r-1}.
\label{eq:D}
\end{equation}
The reader may refer to Example  B in Section IX for a numerical example of the structures $\mathsf{D}$ and $\mathsf{L}$. The goal is to infer the logical dynamical equations of the system from the observed data $\mathsf{D}$, which best explain its behavior and are capable of predicting its evolution under different initial 
states of its nodes. This reverse engineering process relies on the acquisition of sufficient data for the construction of an accurate model. However, it is not always feasible to capture sufficient data and it can also sometimes be very expensive. 

In this article, we propose the Stochastic Conjunctive Normal Form Network (SCNFN) which is equivalent to a PBN but effectively intertwines logic rules and probabilities. 
The SCNFN learning results in a statistical process which can be viewed as a ``logic`` regression problem; it estimates through boolean relationships the entries of the transition probability matrix (dependent variable) while a boolean representation is used for the independent variables (the states $\boldsymbol{\mu}, \boldsymbol{\lambda}$). This process entails a significantly smaller number of parameters that have to be estimated. Therefore, in contrast to the methodologies introduced in prior work, SCNFs can be learned for large networks, from small training datasets.
\section{Proposed Model}
In this section, we introduce the Stochastic Conjunctive Normal Form Network (SCNFN) (Subsection IV.A) and show that it is equivalent to the Probabilistic Boolean Network (PBN) (Subsection IV.B). In the SCNF network, the rule that corresponds to each node (a SCNF formula) consists of a conjunction (logical AND) of multiple disjunctions (logical OR) of boolean variables. Stochasticity is induced at the level of each separate disjunction which is associated with a probability of being activated (evaluated), and in which one literal is actually a Bernoulli random variable.
\subsection{Model Definition}
\begin{definition}
\setstretch{1.35}
A Stochastic Conjunctive Normal Form Network (SCNFN) is a directed network with $N$ binary-valued nodes $V=\{x_1, x_2, \dots, x_N\}$.
Each node $x_i$ has $N^{(i)}$ parent nodes $V^{(i)}=\big\{x^{(i)}_1, x^{(i)}_2, \dots, x^{(i)}_{N^{(i)}}\big\}$. Let $s_t(x) \in \big\{False, True\big\}$ be the 
state of node $x$ at time $t$. Define $s^{(i)}_t \triangleq s_t(x_i)$. The state of node $x_i$ is determined by a Stochastic Conjunctive Normal Form (SCNF) expression  
$\tilde{\Psi}^{(i)}: \{False, True\}^{N^{(i)}} \rightarrow \{False, True\}$ such that: $\tilde{\Psi}^{(i)}=\tilde{\Psi}^{(i)}\big(x^{(i)}_{1}, x^{(i)}_{2},\dots,x^{(i)}_{N^{(i)}}; \mathbf{p}^{(i)}\big)$ with $\mathbf{p}^{(i)}=[p^{(i)}_1, p^{(i)}_2,\dots,p^{(i)}_{M^{(i)}}] \in [0,1]^N$ and $V^{(i)}=\big\{x^{(i)}_{1}, x^{(i)}_{2},\dots,x^{(i)}_{N^{(i)}}\big\} \subseteq V$, the sets of the $N^{(i)}$ parents of node $x_i$.
\allowdisplaybreaks
\begin{align}
& \tilde{\Psi}^{(i)}\big(x^{(i)}_{1}, x^{(i)}_{2},\dots,x^{(i)}_{N^{(i)}}; \mathbf{p}^{(i)}\big)= \nonumber \\
& \qquad \ \ \tilde{\psi}^{(i)}_1\big(x^{(i)}_{1,1}, x^{(i)}_{1, 2},\dots,x^{(i)}_{1,N^{(i)}_1}; p^{(i)}_1\big)  \nonumber \\
& \qquad {\land}\:\tilde{\psi}^{(i)}_2\big(x^{(i)}_{2,1}, x^{(i)}_{2, 2},\dots,x^{(i)}_{2,N^{(i)}_2}; p^{(i)}_2\big) \nonumber\\
& \qquad\qquad\qquad\qquad\hdots\qquad\qquad\qquad \nonumber\\
& \qquad {\land}\:\tilde{\psi}^{(i)}_{M^{(i)}}\big(x^{(i)}_{M^{(i)},1}, x^{(i)}_{M^{(i)}, 2},\dots,x^{(i)}_{M^{(i)},N^{(i)}_{M^{(i)}}}; p^{(i)}_{M^{(i)}}\big).
\end{align}
Let $V^{(i)}_j=\big\{x^{(i)}_{j,1}, x^{(i)}_{j, 2},\dots,x^{(i)}_{j,N^{(i)}_j}\big\}$ be the set of the $N^{(i)}_j$ variables of the clause $\tilde{\psi}^{(i)}_j$. Then $V^{(i)}=\bigcup_{j=1}^{M^{(i)}}V^{(i)}_j$, and $V^{(i)}\subseteq V$. Each clause $\tilde{\psi}^{(i)}_j$ in the conjunction is defined as follows:
\begin{align}
& \tilde{\psi}^{(i)}_{j}\big(x^{(i)}_{j,1}, x^{(i)}_{j, 2},\dots,x^{(i)}_{j,N^{(i)}_j}; p^{(i)}_{j}\big)= \\
& \qquad\psi^{(i)}_j\big(x^{(i)}_{j,1}, x^{(i)}_{j, 2},\dots,x^{(i)}_{j,N^{(i)}_j}\big) \lor \neg \alpha^{(i)}_j,
\end{align}
{where it holds that:}
\begin{align}
& \psi^{(i)}_j\big(x^{(i)}_{j,1}, x^{(i)}_{j, 2},\dots,x^{(i)}_{j,N^{(i)}_j}\big)= \nonumber \\
& \qquad l\big(x^{(i)}_{j,1}\big) \lor \dots \lor l\big(x^{(i)}_{j,N^{(i)}_j}\big), \\
& l\big(x^{(i)}_{j,k}\big) \in \big\{x^{(i)}_{j,k}, \neg x^{(i)}_{j,k}\big\}, \\
& \alpha^{(i)}_j \sim Bernoulli\big(p^{(i)}_j\big), 
\end{align}for $i=1,2,\dots,N$, $j=1,2,\dots,M^{(i)}$ and $k=1,2,\dots, N^{(i)}_j$. The state of the whole network at time step $t$ is represented by the vector $\mathbf{S}_t=\big[s_t^{(1)},s_t^{(2)},\dots, s_t^{(N)}\big] \in \{False, True\}^N$. Let $\tilde{\boldsymbol{\Psi}}=\big[\tilde{\Psi}^{(1)}, \tilde{\Psi}^{(2)},\dots, \tilde{\Psi}^{(N)}\big]$ be the network function. The states of all nodes are updated synchronously (at the same time) and independently. Then, the dynamics after one system transition of the SCNFN is given by: 
\allowdisplaybreaks
\begin{align}
& \mathbf{S}_{t+1}=\tilde{\boldsymbol{\Psi}}(\mathbf{S}_t), \nonumber \\
&s_{t+1}^{(i)}=\tilde{\Psi}^{(i)}(\mathbf{S}_t), \nonumber \\
&\tilde{\Psi}^{(i)}(\mathbf{S}_t)=\tilde{\psi}^{(i)}_1(\mathbf{S}_t) \land \tilde{\psi}^{(i)}_2(\mathbf{S}_t) \land \dots  \land \tilde{\psi}^{(i)}_{M^{(i)}}(\mathbf{S}_t), \nonumber \\
&\tilde{\psi}^{(i)}_j(\mathbf{S}_t)= l\big(\mathbf{s}_t\big(x^{(i)}_{j,1}\big)\big) \lor  l\big(\mathbf{s}_t\big(x^{(i)}_{j,2}\big)\big) \lor \dots  l\big(\mathbf{s}_t\big(x^{(i)}_{j,N^{(i)}_j}\big)\big), \nonumber \\
\allowdisplaybreaks
& l\big(\mathbf{s}_t\big(x^{(i)}_{j,k}\big)\big) = \begin{cases}
\mathbf{s}_t\big(x^{(i)}_{j,k}\big) &\text{ if } l\big(x^{(i)}_{j,k}\big)=x^{(i)}_{j,k} \\
\neg \mathbf{s}_t\big(x^{(i)}_{j,k}\big) &\text{ if } l\big(x^{(i)}_{j,k}\big)=\neg x^{(i)}_{j,k},
\end{cases} \nonumber \\
\label{eq:scnf_dynamics}
\end{align}
for $i=1,2,\dots,N$, $j=1,2,\dots,M^{(i)}$ and $k=1,2,\dots, N^{(i)}_j$, where $\neg$ refers to the logical negation. Similarly, the dynamics after $k$ system transitions is represented as $\mathbf{S}_{t+k}=\tilde{\boldsymbol{\Psi}}^{k}(\mathbf{S}_t)$, where the operator $\tilde{\boldsymbol{\Psi}}^{k}$ corresponds to $k$ repetitions of the update rules in Equation \ref{eq:scnf_dynamics}. We denote the SCNFN by $G(V,\tilde{\boldsymbol{\Psi}})$.
\end{definition} 

By definition, the presence of the Bernoulli random literal $\alpha^{(i)}_j$ in the disjunction $\tilde{\psi}^{(i)}_j$ implies that, with probability $1-p^{(i)}_j$, $\tilde{\psi}^{(i)}_j$  does not contribute to the logical value of $\tilde{{\Psi}}^{(i)}$. That is because:
\begin{equation}
\label{eq:stochastic_disjunction}
\tilde{\psi}^{(i)}_j\big(\mathbf{S}_t; p^{(i)}_j\big) = \begin{cases}
True &\text{ w.p. } \big(1-p^{(i)}_j\big) \\
\psi^{(i)}_j\left(\mathbf{S}_t\right) &\text{ w.p. } \ p^{(i)}_j,
\end{cases}
\end{equation}
where "w.p." stands for "with probability". We will use upper case Greek letters $\tilde{\Theta}^{(i)}, \tilde{\Phi}^{(i)}, \tilde{\Psi}^{(i)}$ to represent the SCNF related to node $i$, and lower case Greek letters $\tilde{\theta}^{(i)}_j, \tilde{\phi}^{(i)}_j, \tilde{\psi}^{(i)}_j$ to represent the $j$th-stochastic disjunction in the SCNF rule of node $i$. Finally, $l^{(i)}_{j,k}$ refers to the $k$-th literal in the $j$-th disjunction in the SCNF rule of node $i$. Without loss of generality, the quantifier $k$ of literal $l^{(i)}_{j,k}$ refers to its lexicographic order within the disjunction $\tilde{\psi}^{(i)}_j$, such that $\neg{x_j} \prec x_j$, and $\neg{x_j} \prec \neg{x_{j'}}$, \eapcomment{$\neg{x_j} \prec x_{j'}$, ${x_j} \prec \neg{x_{j'}}$, ${x_j} \prec x_{j'}$} if $j<j'$ , \iapcomment{while} the quantifier $j$ of the disjunction $\tilde{\psi}^{(i)}_j$ refers to its lexicographic order in the SCNF rule $\tilde{\Psi}^{(i)}$. Note that in the rest of the paper we may equivalently represent a conjunction $\tilde{\Theta}$ as a set of disjunctions, and a disjunction $\tilde{\theta}$ as a set of literals. Therefore,  $\big|\tilde{\Theta}\big|$ is the number of disjunctions in the SCNF rule $\tilde{\Theta}$ and $\big|\tilde{\theta}\big|$ is the number of literals (excluding the Bernoulli random variable) in the stochastic disjunction $\tilde{\theta}$. Finally, we may drop the arguments in a stochastic disjunction or a SCNF which correspond to its variables, or the Bernoulli parameter in case they do not contribute to the understanding of the concepts elaborated.
\subsection{Equivalence of SCNFN and PBN}
We now prove that any SCNFN can be converted into an equivalent PBN and vice-versa. This is expected since they both represent a \iapcomment{discrete-time} homogeneous Markov Chain. This equivalence can also be viewed as the stochastic extension of the conversion of any propositional formula to conjunctive normal form (CNF).
\begin{proposition}
Every SCNFN $G(V,\tilde{\boldsymbol{\Psi}})$ can be converted to a PBN $G(V',\mathbf{\tilde{F}})$.
\end{proposition}
\begin{IEEEproof}
Clearly, $V'=V$. Recall that $\mathbf{\tilde{F}}=(\mathbf{F},\mathbf{P})$. Fix node $i$. We describe the conversion of the SCNF rule $\tilde{\Psi}^{(i)}$ to the vector $\mathbf{f}^{(i)}\iapcomment{=\mathbf{F}(i)}$ and the corresponding selection probabilities $\mathbf{p}^{(i)}\iapcomment{=\mathbf{P}(i)}$. First, assume that $\tilde{\Psi}^{(i)}=\Phi^{(i)} \land \tilde{\Theta}^{(i)}$ is the SCNF rule of node $i$ decomposed in the deterministic portion $\Phi^{(i)}$ and the stochastic portion $\tilde{\Theta}^{(i)}$, such that each disjunction $\phi^{(i)}_j \in \Phi^{(i)}$ is deterministic (i.e., $p^{(i)}_j=1.0$ for $j=1,2,\dots,\big|{\Phi}^{(i)}\big|$). By definition of the activation of the disjunctions in $\tilde{\Theta}^{(i)}$, each rule $f^{(i)}_{j}=\iapcomment{\mathbf{f}^{(i)}}(j)$ will correspond to the logical AND of $\Phi^{(i)}$ and an element of the power set $\mathscr{P}\big(\tilde{\Theta}^{(i)}\big)$ (which contains all possible subsets of the stochastic disjunctions in $\tilde{\Theta}^{(i)}$). Let $\beta_{j}$ be the $j$-th element in $\mathscr{P}\big(\tilde{\Theta}^{(i)}\big)$ for $j=1,2,\dots,2^{\big|\tilde{\Theta}^{(i)}\big|}$ (by assuming lexicographic order of the disjunctions in $\beta, \beta'$\iapcomment{} and $\beta < \beta'$ if $|\beta|<|\beta'|$). Then, the logic rule $f^{(i)}_{j}$ \iapcomment{for $j=1,2,\dots,2^{\big|\tilde{\Theta}^{(i)}\big|}$, } is:
\begin{align}
& f^{(i)}_{j}=\Phi^{(i)} \land \mathscr{Beta}_{j}\big(\tilde{\theta}^{(i)}_1\big) \land \dots \land \mathscr{Beta}_{j}\big(\tilde{\theta}^{(i)}_{|\tilde{\Theta}^{(i)}|}\big),\nonumber \\
& \mathscr{Beta}_{j}\big(\tilde{\theta}^{(i)}_z\big)= \begin{cases}
\tilde{\theta}^{(i)}_z &\text{ if } \tilde{\theta}^{(i)}_z \in \beta_{j} \\
True &\text{ if } \tilde{\theta}^{(i)}_z \notin \beta_{j},
\end{cases}
\end{align}
for $z=1,2,\dots,\big|\tilde{\Theta}^{(i)}\big|$, while the corresponding selection probability $p^{(i)}_{j} \in \mathbf{p}^{(i)}(j)$  will be:
\begin{align}
p^{(i)}_{j}=\mathlarger{\prod}\limits_{z=1}^{|\tilde{\Theta}^{(i)}|} {p^{(i)}_z}^{I\big(\tilde{\theta}^{(i)}_z \in \iapcomment{\beta_{j}}\big)}\big(1-p^{(i)}_z\big)^{I\big(\tilde{\theta}^{(i)}_z \notin \iapcomment{\beta_{j}}\big)}, 
\end{align}
where $I$ the numerical indicator function which returns 1 if the condition in its argument is $True$, otherwise it returns 0.
The above formulation constitutes a PBN.
\end{IEEEproof}
\begin{proposition}
Every PBN $G(V,\mathbf{\tilde{F}})$ can be converted to a SCNFN $ G(V',\tilde{\boldsymbol{\Psi}})$.
\end{proposition}
\begin{IEEEproof}
Clearly $V'=V$. Recall that $\mathbf{\tilde{F}}=(\mathbf{F},\mathbf{P})$. Fix node $i$. We will describe the conversion of the vector $\mathbf{f}^{(i)}\iapcomment{=\mathbf{F}(i)}$ which contains all the $M^{(i)}$ logic rules which regulate the dynamics of node $i$ and the corresponding selection probabilities $\mathbf{p}^{(i)}\iapcomment{=\mathbf{P}(i)}$ to the equivalent SCNF rule $\tilde{\Psi}^{(i)}\iapcomment{=\tilde{\boldsymbol{\Psi}}(i)}$. The \iapcomment{SCNF} $\tilde{\Psi}^{(i)}$ consists of $M'^{(i)}=2^N$ stochastic disjunctions $\tilde{\psi}^{(i)}_j$ such that:
\begin{equation}
\label{eq:h_x}
\tilde{\Psi}^{(i)} = \tilde{\psi}^{(i)}_0 \land \tilde{\psi}^{(i)}_1 \land \dots \land \tilde{\psi}^{(i)}_{2^N-1}.
\end{equation}
Each disjunction:
\begin{equation}
\tilde{\psi}^{(i)}_j(x_1, \dots, x_N; q^{(i)}_j)=\psi^{(i)}_j(x_1, \dots, x_N; q^{(i)}_j) \lor \alpha^{(i)}_j,
\end{equation} for $j=0,1,\dots,2^N-1$, corresponds to each possible combination of the available literals $L=\{x_i, \neg{x_i}\}_{i=1}^{N}$ and can become $False$ for exactly one network state. 
Therefore, 
\allowdisplaybreaks
\begin{align}
\label{eq:d_x}
& \psi^{(i)}_0 = (\neg{x_1} \lor \dots \lor \neg{x_{2^N-1}} \lor \neg{x_{2^N}}), \nonumber \\
& \psi^{(i)}_1 = (\neg{x_1} \lor \dots \lor \neg{x_{2^N-1}} \lor {x_{2^N}}), \nonumber \\
& \psi^{(i)}_2 = (\neg{x_1} \lor \dots \lor {x_{2^N-1}} \lor \neg{x_{2^N}}), \nonumber \\
& \psi^{(i)}_3 = (\neg{x_1} \lor \dots \lor {x_{2^N-1}} \lor {x_{2^N}}), \nonumber \\
& \qquad\qquad\qquad\hdots\qquad\qquad\qquad \nonumber \\
& \psi^{(i)}_{2^N-1} = ({x_1} \lor \dots \lor {x_{2^N-1}} \lor {x_{2^N}}).
\end{align}

In order to find the parameter $q^{(i)}_j$ of the stochastic disjunction $\tilde{\psi}^{(i)}_j$, we should first find the unique state $\iapcomment{\boldsymbol{\lambda}^{(i)}_j} \in \{False, True\}^N$, which can yield $False$ when the disjunction, and therefore the full CNF, is evaluated. Note that $\iapcomment{q^{(i)}_j}$ should be equal to the probability that the SCNF $\tilde{\Psi}^{(i)}$  will be evaluated as $False$ for the state $\iapcomment{\boldsymbol{\lambda}^{(i)}_j}$, because the rest of the disjunctions will always evaluate to $True$ (no matter what the outcome of their associated Bernoulli variable\iapcomment{s} is) and, therefore, they have no effect on the evaluation of the system for the state $\iapcomment{\boldsymbol{\lambda}^{(i)}_j}$. 

Subsequently, we compute the probability that the PBN will be evaluated as $True$ for the state $\boldsymbol{\lambda}^{(i)}_j$, given the constituent Boolean rules $f^{(i)}_j$ and their corresponding selection probabilities $p^{(i)}_j$. Therefore, the parameter of each Bernoulli variable $q^{(i)}_j$ of the $j$-th stochastic disjunction, can be described by the formula:
\begin{equation}
\label{eq:q_i_j}
q^{(i)}_j=1- \sum\limits_{j=1}^{M^{(i)}}{p^{(i)}_j I\big(f^{(i)}_j\big(\iapcomment{\boldsymbol{\lambda}^{(i)}_j}\big)\big)},
\end{equation}
where $I$ is the indicator function (which returns 1 if the condition of its argument is $True$).
\end{IEEEproof}
\section{Learning of the SCNF Network}
\subsection{Overview of the Approach}
We now describe the general idea and the main components involved in learning a SCNF network. 
The algorithm greedily, and not optimally, maximizes the likelihood of the time series used \iapcomment{in training}. Finding the optimal solution is a problem of combinatorial complexity, a fact which prevents any exact algorithm to be applicable to large-scale structures. 
The reconstruction methodology consists of two parts:
\begin{enumerate}
\item learning the logical interactions between the nodes in the system (Algorithm \ref{algo:scnf_logic_learn} and Algorithm \ref{algo:scnf_disjunction_learn})\iapcomment{.}
\item learning the parameters of the Bernoulli random variables associated with the disjunctions \iapcomment{discovered} in the previous step (Algorithm \ref{algo:scnf_param_learn}).
\end{enumerate}
\begin{algorithm}
\caption {$\textit{SCNFN-Learn}$}
\label{algo:scnf_network_learn}
\begin{algorithmic}[1]
\scriptsize
\State \textbf{Inputs} 
\State \ \ \ $N$: \iapcomment{The} number of nodes in the system.
\State {\ \ \ $\mathsf{L}$: \iapcomment{The} list with system transitions (Equation \ref{eq:D}).} 
\State \textbf{Output} 
\State \ \ \ $\tilde{\boldsymbol{\Psi}}$: The set of SCNF rules for the system.

\State \textbf{Begin}  \newline

\State $L=\{x_i, \neg x_i\}_{i=1}^{N}$



\State $\tilde{\boldsymbol{\Psi}} \leftarrow []$
\State  \textbf{For }$i=1,2,\dots, N$ 

\State {\ \ \  Form $\mathsf{L}^{(i)}$ (Equation \ref{eq:D_i}).}
\State { \ \ \ $\tilde{\Psi}^{(i)}\leftarrow SCNF-Learn\big(\mathsf{L}^{(i)},L\big)$}
\State { \ \ \ $\tilde{\boldsymbol{\Psi}}(i)\leftarrow \tilde{\Psi}^{(i)}$}
\State \textbf{EndFor}
\newline
\State \textbf{Return} $\tilde{\boldsymbol{\Psi}}$
\end{algorithmic}
\end{algorithm}
Each iteration of Algorithm \ref{algo:scnf_network_learn} learns the SCNF clause of a node $i$ (Algorithm 1, Line 11). Given the list $\mathsf{L}$ (Equation \ref{eq:D}), the reduced list $\mathsf{L}^{(i)}$, which holds pairs of the previous system state and node i's next state is formed, as follows:
\begin{equation}
\mathsf{L}^{(i)}\triangleq\left(\big(\mathbf{S},s'^{(i)}\big): \big(\mathbf{S},\mathbf{S}'\big) \in \mathsf{L}\right).
\label{eq:D_i}
\end{equation}
In the preprocessing step of Algorithm \ref{algo:scnf_learn}, the structure $\mathsf{L}^{(i)}$ is  parsed and the sets
$\mathsf{S}^{(i)}_F$, $\mathsf{S}^{(i)}_T$, $\mathsf{S}^{(i)}_C$ are formed. 
The set $\mathsf{S}^{(i)}_F$ (Algorithm \ref{algo:scnf_learn}, Line 7) consists of the states in \iapcomment{$\mathsf{L}^{(i)}$} which yield only $False$ for the node $i$ as a next state:
\begin{equation}
\mathsf{S}^{(i)}_F\triangleq\big\{ \mathbf{S}: (\mathbf{S},False) \in \mathsf{L}^{(i)}, (\mathbf{S},True) \notin \mathsf{L}^{(i)}  \big\}.
\label{eq:D_i_F}
\end{equation}
Similarly, the set $\mathsf{S}^{(i)}_T$ (Algorithm \ref{algo:scnf_learn}, Line 8) consists of the states in 
$\mathsf{L}^{(i)}$ which yield only $True$ for the node $i$ as a next state:
\begin{equation}
\mathsf{S}^{(i)}_T\triangleq\big\{ \mathbf{S}: (\mathbf{S},True) \in \mathsf{L}^{(i)}, (\mathbf{S},False) \notin \mathsf{L}^{(i)}  \big\}.
\label{eq:D_i_T}
\end{equation}
Finally, the set $\mathsf{S}^{(i)}_C$ contains the system states which drive node $i$ both to a $False$ and a $True$ state in $\mathsf{L}^{(i)}$:
\begin{equation}
\mathsf{S}^{(i)}_C\triangleq\big\{ \mathbf{S}: (\mathbf{S},True) \in \mathsf{L}^{(i)}, (\mathbf{S},False) \in \mathsf{L}^{(i)}  \big\}.
\label{eq:D_i_C}
\end{equation}
\begin{algorithm}
\caption {$\textit{SCNF-Learn}$}
\label{algo:scnf_learn}
\begin{algorithmic}[1]
\scriptsize
\State \textbf{Inputs}
\State \ \ \ $\mathsf{L}^{(i)}$: The list of transitions of a node $i$ (Equation \ref{eq:D_i}).
\State \ \ \ $L$: The set of the available literals.

\State \textbf{Output} 
\State \ \ \ $\tilde{\Psi}^{(i)}$: \iapcomment{The} SCNF formula of the node $i$. 

\State \textbf{Begin} \newline 

\State {\ \ \ Form $\mathsf{S}^{(i)}_F$ (Equation \ref{eq:D_i_F}).}
\State {\ \ \ Form $\mathsf{S}^{(i)}_T$ (Equation \ref{eq:D_i_T}).}
\State {\ \ \ Form $\mathsf{S}^{(i)}_C$ (Equation \ref{eq:D_i_C}).}\newline
\State {\ \ \  $\Phi^{(i)}\big(V^{(i)}_d\big)\leftarrow  CNF-LogicLearn\big(\mathsf{S}^{(i)}_F, \mathsf{S}^{(i)}_T \cup \mathsf{S}^{(i)}_C, L\big)$}\newline
\State {\ \ \  $\tilde{\Theta}^{(i)}\big(V^{(i)}_s\big)\leftarrow  CNF-LogicLearn\big(\mathsf{S}^{(i)}_C, \mathsf{S}^{(i)}_T , L\big)$}\newline
%
%
\State {\ \ \ $\mathbf{p}^{(i)} \leftarrow  CNF-ParameterLearn\big(
\mathsf{L}^{(i)}, \mathsf{S}^{(i)}_C,\tilde{\Theta}^{(i)}\big)$}\newline

\State {\ \ \ $\tilde{\Psi}^{(i)}\big(V^{(i)}_d \cup V^{(i)}_s; \mathbf{p}^{(i)}\big) \leftarrow \Phi^{(i)}\big(V^{(i)}_d\big) \land \tilde{\Theta}^{(i)}\big(V^{(i)}_s;  \mathbf{p}^{(i)}\big) $}\newline

\State \textbf{Return} $\tilde{\Psi}^{(i)}$
\end{algorithmic}
\end{algorithm}

The learning of the SCNF formula (Algorithm \ref{algo:scnf_learn}) involves three steps  (Lines 10, 11, 12 in Algorithm \ref{algo:scnf_learn}). Initially, the deterministic logic portion $\Phi^{(i)}$, where all the disjunctions are always evaluated (the corresponding parameter is 1.0) of the SCNF rule is learned (Line 10 of Algorithm \ref{algo:scnf_learn}). The function \textit{CNF-LogicLearn} (Algorithm \ref{algo:scnf_logic_learn}) is responsible for returning a CNF rule which gives $False$ for all the transitions passed in its first argument and $True$ for all the transitions contained in the second argument. Note here that the transitions in $\mathsf{S}^{(i)}_C$ are treated as positive, since we want to avoid them being evaluated always as $False$ by $\Phi^{(i)}$. 

Afterwards, the stochastic part $\tilde{\Theta}^{(i)}$ is learned (Line 11 of Algorithm \ref{algo:scnf_learn}). The transitions in $\mathsf{S}^{(i)}_F$ can now be ignored, because there exists at least one disjunction in $\Phi^{(i)}$ which turns to $False$ (so that the whole $\Phi^{(i)}$ turns to $False$), when the previous state in $\mathsf{S}^{(i)}_F$ is substituted in its variables $V^{(i)}_d$. On the other hand, the transitions in $\mathsf{S}^{(i)}_C$ are now treated as negative by $\tilde{\Theta}^{(i)}$ since there should be at least one disjunction which probabilistically turns to $False$ when its previous state in $\mathsf{S}^{(i)}_C$ is plugged in its variables $V^{(i)}_s$; otherwise the transition would have been deterministically evaluated as $True$.  The transitions in $\mathsf{S}^{(i)}_T$ should be treated as positive by $\tilde{\Theta}^{(i)}$ as well so that the concatenated $\tilde{\Psi}^{(i)}=\Phi^{(i)} \land \tilde{\Theta}^{(i)}$ is deterministically evaluated as True. 
Once the learning of the logic parts has finished, the algorithm proceeds to learning the parametric part of $\tilde{\Theta}^{(i)}$ (Line 12 in Algorithm \ref{algo:scnf_learn}). Finally, the CNF $\Phi^{(i)}$ and the SCNF $\tilde{\Theta}^{(i)}$ \iapcomment{are} merged into the full SCNF rule $\tilde{\Psi}^{(i)}$ of node $i$ (Line \iapcomment{13} in Algorithm \ref{algo:scnf_learn}).
\subsection{Logic Learning}
In this subsection, we will delve into the details of learning the logic rules (Algorithms \ref{algo:scnf_logic_learn}, \ref{algo:scnf_disjunction_learn}). Algorithm \ref{algo:scnf_logic_learn} returns a CNF rule which is $False$ for all the states in $H_F$ (Line 6.i) and $True$ for all the states in $H_T$ (Line 6.ii). This implies that each disjunction in $\Phi$ should be $True$ for all states in $H_T$ and that for each state in $H_F$, there should be at least one disjunction in $\Phi$ which is $False$. Each loop iteration in Algorithm \ref{algo:scnf_logic_learn} adds a new disjunction $\phi_k$ in the currently formed conjunction $\Phi$ (Line 11). The function $Disjunction-Learn$ is responsible for returning a disjunction which satisfies two conditions: i) it evaluates as $True$ all the states passed in its first argument ii) it evaluates as $False$ at least one state passed in its second argument. In this way, it is guaranteed that the loop in Lines 10-15 of Algorithm \ref{algo:scnf_logic_learn} will not perpetually add disjunctions and therefore the condition in Line 15 will finally become $True$. Consequently, the completeness of Algorithm \ref{algo:scnf_logic_learn} is guaranteed. 
\begin{algorithm}
\caption {$\textit{CNF-LogicLearn}$}
\label{algo:scnf_logic_learn}
\begin{algorithmic}[1]
\scriptsize
\State \textbf{Inputs}
\State \ \ \ $H_F$: The set of negative transitions.
\State \ \ \ $H_T$: The set of positive transitions.
\State \ \ \ $L$: The set of the available literals.
\State \textbf{Output} 
\State {\ \ \ $\Phi$: a CNF clause, which satisfies: 

 i) $\Phi(\mathbf{S})=False, \forall \mathbf{S} \in H_F$ 
 
 ii)\iapcomment{  }$\Phi(\mathbf{S})=True, \forall \mathbf{S} \in H_T$}
\State \textbf{Begin}  \newline

\State  $k \leftarrow 0 $
\State  $\Phi \leftarrow \emptyset $
\State  \textbf{Repeat}
\State {\ \ \ $\phi_k \leftarrow  Disjunction-Learn(H_F, H_T, L,\emptyset)$}
\State {\ \ \ $H_F \leftarrow H_F-\{\mathbf{S}: \mathbf{S} \in H_F, \phi_k(\mathbf{S})=False $\} }
\State {\ \ \  $\Phi \leftarrow \Phi \land \phi_k
$}
\State {\ \ \  $k \leftarrow k+1 $}
\State \textbf{Until}($H_F==\emptyset$)\newline

\State \textbf{Return} $\Phi$
\end{algorithmic}
\end{algorithm}
In case a transition in $H_F$ is evaluated as $False$ by the disjunction $\phi_k$ it is removed (Line 12), since it suffices if there exists at least one disjunction which turns to False, so that the whole $\Phi$ yields $False$. Note that $H_T$ does not change since each disjunction $\phi_k$ should be evaluated as $True$ for each transition in $H_T$ so that the final CNF clause $\Phi$ is in accordance with both $H_F$ and $H_T$. Once $H_F$ becomes empty (all the negative transitions are satisfied by $\Phi$) (Line 15, Algorithm \ref{algo:scnf_logic_learn}), the function returns. One recursive implementation of the \iapcomment{auxiliary }function $Disjunction-Learn$ accompanied by explanatory comments is provided in Algorithm $\ref{algo:scnf_disjunction_learn}$ in the Appendix.
\subsection{Parameter Learning}
At the last step of the learning procedure, the parameters \iapcomment{${p}^{(i)}_j$} of the stochastic disjunctions $\iapcomment{\tilde{\psi}^{(i)}_j}$ in the SCNF rule of the node $i$, should be estimated so that the likelihood of the time series (transitions in $\mathsf{L}^{(i)}$, Equation \ref{eq:D_i}) is maximized (Algorithm \ref{algo:scnf_param_learn}). Towards this goal, some structures have to be defined.

Let $\tilde{\Psi}^{(i)}_F(\boldsymbol{\lambda})$ be the set of stochastic disjunctions in $\tilde{\Psi}^{(i)}$ which can be evaluated as $False$ (if they are activated; i.e., $\alpha^{(i)}_j=True$) for the state $\boldsymbol{\lambda}$: 
\begin{equation}
\tilde{\Psi}^{(i)}_F(\boldsymbol{\lambda})\triangleq\left\{ \tilde{\psi}^{(i)}_j: \tilde{\psi}^{(i)}_j \in \tilde{\Psi}^{(i)}, \tilde{\psi}^{(i)}_j\big(\boldsymbol{\lambda}\big)=False\right\}.
\label{eq:Psi_i_F_lambda}
\end{equation}
\iapcomment{Note that the omission of $p^{(i)}_j$ in the arguments of $\tilde{\psi}^{(i)}_j$ above implies that we do care about the value of $\tilde{\psi}^{(i)}_j$ in case $\alpha^{(i)}_j=True$.}

Let $\mathscr{P}_F^{(i)}\big(\boldsymbol{\lambda}\big)$ represent the probability that the previous state $\boldsymbol{\lambda}$ of the system will turn the node $i$ to $False$ in the next time step. Formally,
\begin{equation}
\mathscr{P}_F^{(i)}\left(\boldsymbol{\lambda}\right)\triangleq Prob\big(\tilde{\Psi}^{(i)}\big(\boldsymbol{\lambda}; \mathbf{p}^{(i)}\big)=False\big).
\label{eq:P_Psi_F_lambda}
\end{equation}
It can be computed by using the inclusion-exclusion principle \iapcomment{\cite{wasserman2013all}} on the activation of at least one disjunction in $\tilde{\Psi}^{(i)}_F:$
\begin{equation}
 \label{P_F_lambda} 
\mathscr{P}_F^{(i)}\big(\boldsymbol{\lambda}\big)=\mathlarger{\sum}\limits_{m=1}^{|\tilde{\Psi}^{(i)}_F(\boldsymbol{\lambda})|} \big((-1)^{(m+1)}\sum\limits_{\forall \tilde{\Theta}_{j_m}\in \binom{\tilde{\Psi}^{(i)}_F(\boldsymbol{\lambda})}{m}}^{} \prod\limits_{k=1}^{m}p^{(i)}_{j_m(k)}
 \big).
 \end{equation}
In the above equation, $\binom{S}{m}$ is the set of all subsets of the set $S$ of cardinality $m$. $\tilde{\Theta}_{j_m}$ for $j_m=1, 2,\dots, \big|\binom{\tilde{\Psi}^{(i)}_F(\boldsymbol{\lambda})}{m}\big|$,  denotes the $j_m$-th element/SCNF in $\binom{\tilde{\Psi}^{(i)}_F(\boldsymbol{\lambda})}{m}$ (by assuming lexicographic order \iapcomment{of the SCNFs} as before) such that:
\begin{equation}
\tilde{\Theta}_{j_m}=\tilde{\psi}^{(i)}_{j_m(1)} \land \tilde{\psi}^{(i)}_{j_m(2)} \land \dots \land \tilde{\psi}^{(i)}_{j_m(m)},
\end{equation}
\iapcomment{with $\tilde{\psi}^{(i)}_{j_m(k)} \in \tilde{\Psi}^{(i)}_F(\boldsymbol{\lambda})$ denoting the $k$-th stochastic disjunction in the $j_m$-th SCNF assuming a lexicographic order of the disjunctions as before.}

We should also define $\mathscr{N}^{(i)}_F(\boldsymbol{\lambda})$, which is the number of times the system state $\boldsymbol{\lambda}$ drives node $i$ to a low state in the list $\mathsf{L}^{(i)}$:
\begin{equation}
\mathscr{N}^{(i)}_F(\boldsymbol{\lambda})\triangleq\left|\big(\mathbf{S}: (\mathbf{S},s) \in \mathsf{L}^{(i)}, \mathbf{S}=\boldsymbol{\lambda}, s=False \big)\right|.
\label{eq:D_i_F_lambda}
\end{equation}
Similarly, $\mathscr{N}^{(i)}_T(\boldsymbol{\lambda})$ can be defined as follows: 
\begin{equation}
\mathscr{N}^{(i)}_T(\boldsymbol{\lambda})\triangleq\left|\big(\mathbf{S}: (\mathbf{S},s) \in \mathsf{L}^{(i)}, \mathbf{S}=\boldsymbol{\lambda}, s=True \big)\right|.
\label{eq:D_i_T_lambda}
\end{equation}

Given the definitions of all the aforementioned structures, the likelihood of the time
series (the probability of observing the temporal sequence) $\mathsf{L}^{(i)}$ of the node $i$ is given by the  Equation \ref{eq:reduced_L_i}:
\begin{equation}
\label{eq:reduced_L_i}
\mathcal{L}_i\big(\mathsf{L}^{(i)}; \mathbf{p}^{(i)}\big)=\mathlarger{\prod}\limits_{\forall \boldsymbol{\lambda}^{} \in \mathsf{S}^{(i)}_C}^{}\mathscr{P}_F^{(i)}\big(\boldsymbol{\lambda}\big)^{\mathscr{N}^{(i)}_F(\boldsymbol{\lambda})}\big(1-\mathscr{P}_F^{(i)}\big(\boldsymbol{\lambda}\big)\big)^{\mathscr{N}^{(i)}_T(\boldsymbol{\lambda})}
\end{equation}
In order to derive the above formula, we may consider each out of the $\mathscr{N}^{(i)}_F(\boldsymbol{\lambda})+\mathscr{N}^{(i)}_T(\boldsymbol{\lambda})$ appearances of the state $\boldsymbol{\lambda}$ as a realization of a binomial process which yields $\mathscr{N}^{(i)}_F(\boldsymbol{\lambda})$ times $False$ and $\mathscr{N}^{(i)}_T(\boldsymbol{\lambda})$ $True$ as next state for the node $i$ under consideration.
Moreover, only the transitions in $\mathsf{S}^{(i)}_C$ contribute to the estimation of $\mathbf{p}^{(i)}$ since $\forall \boldsymbol{\lambda} \in \mathsf{S}^{(i)}_F$\iapcomment{:} $\mathscr{P}^{(i)}_F\left( \boldsymbol{\lambda}\right)=1.0$ and $\forall \boldsymbol{\lambda} \in \mathsf{S}^{(i)}_T$\iapcomment{:} $\mathscr{P}^{(i)}_F\left( \boldsymbol{\lambda}\right)=0.0$, due to the constraints in the construction of the logical rules.
Subsequently, by taking the natural logarithm of Equation 
\ref{eq:reduced_L_i}, we obtain Equation \ref{eq:l_i} (the log-likelihood):
\begin{equation}
\label{eq:l_i}
\mathcal{\ell}_i\big(\mathsf{L}^{(i)};\mathbf{p}^{(i)}\big)\triangleq log\big(\mathcal{L}_i\big(\mathsf{L}^{(i)}; \mathbf{p}^{(i)}\big)\big).
\end{equation}

We can now formulate the optimization problem to find the estimate of $\mathbf{p}^{(i)}$ which maximizes the likelihood of the transitions in the training set $\mathsf{L}^{(i)}$ (or equivalently minimizes the negative log-likelihood):
\allowdisplaybreaks
\begin{align}
&\min_{\substack{\mathbf{p}^{(i)} \\ \left\{\mathscr{P}^{(i)}_F\left(\boldsymbol{\lambda}\right) \right\}_{ \forall \boldsymbol{\lambda} \in \mathsf{S}^{(i)}_{C}}}} - \sum\limits_{\forall \boldsymbol{\lambda} \in \mathsf{S}^{(i)}_C }^{} \ell_i(\boldsymbol{\lambda}), 
\label{eq:min_ll}\\
&\text{where: } \nonumber\\ 
& 
\ell_i(\boldsymbol{\lambda})=\mathscr{N}^{(i)}_F(\boldsymbol{\lambda})\log\big(\mathscr{P}_F^{(i)}\big(\boldsymbol{\lambda}\big)\big) + \mathscr{N}^{(i)}_T(\boldsymbol{\lambda})\log\big(1-\mathscr{P}_F^{(i)}\big(\boldsymbol{\lambda}\big)\big), \label{eq:li(s)} 
\\
&\text{subject to: } \nonumber\\
&\mathbf{0} \leq \mathbf{p}^{(i)} \leq \mathbf{1}, \mathbf{p}^{(i)} \in \mathbb{R}^{|\tilde{\Theta}^{(i)}|} \label{eq:pi}\\
&\forall \boldsymbol{\lambda} \in \mathsf{S}^{(i)}_C : \nonumber\\
& 
\mathscr{P}_F^{(i)}\big(\boldsymbol{\lambda}\big)=\mathlarger{\sum}\limits_{m=1}^{|\tilde{\Psi}^{(i)}_F(\boldsymbol{\lambda})|} \big((-1)^{(m+1)}\sum\limits_{\forall \tilde{\Theta}_{j_m}\in \binom{\tilde{\Psi}^{(i)}_F(\boldsymbol{\lambda})}{m}}^{} \prod\limits_{k=1}^{m}p^{(i)}_{j_m(k)}
 \big). 
 \label{eq:pi(s)}
\end{align}

\begin{algorithm}
\caption {$\textit{CNF-ParameterLearn}$}
\label{algo:scnf_param_learn}
\begin{algorithmic}[1]
\scriptsize
\State \textbf{Inputs} 
\State \ \ \ $\mathsf{L}^{(i)}$: The list of transitions of a node $i$.
\State \ \ \ $\mathsf{S}^{(i)}_C$: The set of conflict transitions.
\State \ \ \ $\tilde{\Theta}^{(i)}$: The logic part of the stochastic portion of the SCNF.
\iapcomment{\State \ \ \ $\lambda$: The regularization parameter.}
\State \textbf{Output} 
\State \ \ \ $\mathbf{p}^{(i)}$: The parametric part of the stochastic part of the SCNF.

\State \textbf{Begin}  \newline

\State \ \ \ $\varepsilon_i^2 \leftarrow 0$
\State \ \ \  $\mathcal{\ell}_i \leftarrow 0$

\State \ \ \ \textbf{For} $\forall \iapcomment{\boldsymbol{\lambda}} \in \mathsf{S}^{(i)}_C$ 

\State {\ \ \ \ \ \ Form $\tilde{\Psi}^{(i)}_F(\boldsymbol{\lambda})$ (Equation \ref{eq:Psi_i_F_lambda}}).

\State {\ \ \ \ \ \ Form $\mathscr{P}_F^{(i)}\big( \boldsymbol{\lambda}\big)$ (Equation \ref{P_F_lambda}).}

\State {\ \ \ \ \ \ Form $\varepsilon_i(\boldsymbol{\lambda})$ (Equation \ref{eq:pi(s)_rel}).}

\State {\ \ \ \ \ \ $\varepsilon_i^2 \leftarrow \varepsilon_i^2+ \varepsilon_i^2(\boldsymbol{\lambda}$)
}\newline 

\State {\ \ \ \ \ \ Compute $\mathscr{N}^{(i)}_F(\boldsymbol{\lambda})$ (Equation \ref{eq:D_i_F_lambda})}.

\State {\ \ \ \ \ \ Compute $\mathscr{N}^{(i)}_T(\boldsymbol{\lambda})$ (Equation \ref{eq:D_i_T_lambda})}.

\State {\ \ \ \ \ \ Form   $\ell_i(\boldsymbol{\lambda})$ (Equation \ref{eq:li(s)}).}
\State {\ \ \ \ \ \ $\ell_i \leftarrow \ell_i + \ell_i(\boldsymbol{\lambda})$
}

\State \ \ \ \textbf{EndFor}
\newline
\State \ \ \ Form optimization problem OP (\iapcomment{Equations \ref{eq:min_ll_rel}-\ref{eq:pi(s)_rel}}).
\State \ \ \ $\left(\mathbf{p}^{(i)},  
\left\{\mathscr{P}_F^{(i)}\big(\boldsymbol{\lambda}\big)\right\}_{\forall \boldsymbol{\lambda} \in \mathsf{S}^{(i)}_C}\right) \leftarrow conSolve(OP)$\footnotemark
\newline
\State \textbf{Return} $\mathbf{p}^{(i)}$
\end{algorithmic}
\end{algorithm}
\footnotetext{the routine \iapcomment{$conSolve$ refers to TOMLAB's conSolve solver for general, constrained, nonlinear optimization problems.}}

However, it is very likely that the feasible region of the optimization problem as defined in Equations \ref{eq:min_ll}\iapcomment{-}\ref{eq:pi(s)} is the empty set due to the constraint between the optimization variables $\mathscr{P}_F\big(\tilde{\Psi}^{(i)}, \boldsymbol{\lambda}\big)$ and $\mathbf{p}^{(i)}$ in Equation \ref{eq:pi(s)}. 
By relaxing this constraint, the above problem can be converted as follows (Equations 
\ref{eq:min_ll_rel}\iapcomment{-} \ref{eq:pi(s)_rel}):
\allowdisplaybreaks
\begin{align}
& \min_{\substack{\mathbf{p}^{(i)} \\ \left\{\mathscr{P}^{(i)}_F\left(\boldsymbol{\lambda}\right) \right\}_{ \forall \boldsymbol{\lambda} \in \mathsf{S}^{(i)}_{C}}}} - \sum\limits_{\forall \boldsymbol{\lambda} \in \mathsf{S}^{(i)}_C }^{} \ell_i(\boldsymbol{\lambda}) + \lambda \sum\limits_{\forall \boldsymbol{\lambda} \in \mathsf{S}^{(i)}_C }^{} \varepsilon_i^2(\boldsymbol{\lambda}),
\label{eq:min_ll_rel}\\
&\text{where: } \nonumber\\ 
& \lambda \geq 0 \nonumber \\
& 
\ell_i(\boldsymbol{\lambda})=\mathscr{N}^{(i)}_F(\boldsymbol{\lambda})\log\left(\mathscr{P}_F^{(i)}\big(\boldsymbol{\lambda}\big)\right) + \mathscr{N}^{(i)}_T(\boldsymbol{\lambda})\log\left(1-\mathscr{P}_F^{(i)}\big(\boldsymbol{\lambda}\big)\right) 
, \label{eq:li(s)_rel}\\
&\text{subject to: } \nonumber\\
&\mathbf{0} \leq \mathbf{p}^{(i)} \leq \mathbf{1}, \mathbf{p}^{(i)} \in \mathbb{R}^{|\tilde{\Theta}^{(i)}|}, \label{eq:pi_rel}\\
&\forall \boldsymbol{\lambda} \in \mathsf{S}^{(i)}_C : \nonumber\\
& 
\varepsilon_i(\boldsymbol{\lambda})=\mathscr{P}_F^{(i)}\big(\boldsymbol{\lambda}\big)-\mathlarger{\sum}\limits_{m=1}^{|\tilde{\Psi}^{(i)}_F(\boldsymbol{\lambda})|} \big((-1)^{(m+1)}\sum\limits_{\forall \tilde{\Theta}_{j_m}\in \binom{\tilde{\Psi}^{(i)}_F(\boldsymbol{\lambda})}{m}}^{} \prod\limits_{k=1}^{m}p^{(i)}_{j_m(k)}
 \big). 
 \label{eq:pi(s)_rel}
\end{align}

The aforementioned problem is not convex, therefore it has to be solved for a varying number of regularization parameters $\lambda$  in order to avoid local optima. A similar relaxation of probability constraints was used in \cite{platanios2014estimating}. Our implementation uses the TOMLAB Base Module "conSolve" solver\iapcomment{, which is standard for solving of general, constrained, nonlinear optimization problems.}
\subsection{Comments and Future Extensions}
We should mention that the core of our methodology lies on the fact that the structure learning guides the parameter learning through the disjunctions learned at this step. This way, our heuristic can circumvent the combinatorial nature of the joint structure/parameter learning problem. We should remark here that only the number of different combinations of literals which can participate in the SCNF of a node is $O(2^{2^N})$ \cite{ching2005construction}.  Of course, the structure learning during the first two steps (Algorithm \ref{algo:scnf_learn}, Lines 10, 11) is not optimal, in the sense that a different combination of disjunctions and the corresponding learned parameters could achieve larger value in Equation \ref{eq:reduced_L_i}. However, our simulation results verify that the solution discovered, albeit not optimal, can still achieve very good accuracy. Consequently, the small accuracy loss is very well compensated by the low computational cost (compared to the complexity of learning the globally optimal PBN which maximizes the likelihood of the timeseries). 

\iapcomment{The learning algorithm can be inherently extended to incorporate prior knowledge on network structure, which as suggested in \cite{shmulevich2010probabilistic}, \cite{liu2008inference} can improve the accuracy of the learning. This can be accomplished by modifying the set of the available literals $L$ so that it only includes literals related to nodes with which it is known that a node interacts. A Temporal Boolean Network \cite{silvescu2001temporal}, in which the next state of the system can depend on many states in the past, can also be learned by augmenting the set $L$ so that it includes literals for the temporal delays.}

\iapcomment{Finally, the likelihood in Equation \ref{eq:reduced_L_i} can be modified to express more complex dynamics which can be drawn from asynchronous PBNs or time-dependent transition probabilities, so that the parameters which describe these dependencies are also learned.}
\section{Experimental Results}
In this section, we test the predictive capability of the learning algorithm presented. Before proceeding to the details of the experiments, we would like to point out that the state of the art method \cite{marshall2007inference} manages to reconstruct the PBN of a 7-node network from an unrealistically large amount of training data. More specifically, predictive capacity emerges after training with a single temporal sequence of ~20000 transitions which is much larger than the data complexity required for the reconstruction of the 10-node network with our method. Moreover, a very small maximum node in-degree $k=4$ is assumed, which seriously restricts the class of PBNs, especially for large $N$, that can be reconstructed while no full-dynamics prediction is reported. Therefore, we did not replicate their results. The library in \cite{doi:10.1093/bioinformatics/btq124} purports that it supports learning of a PBN, but in fact no parameters are estimated. If more than one Boolean Network satisfies the time series data, then the candidate Boolean networks are uniformly selected. This approach yields predictive capacity slightly better than chance, therefore we omitted the presentation of these results in this article. As before, a maximum node in-degree is assumed. Their implementation failed to terminate within a reasonable amount of time for $k=5$ and $N=100$. Since then, to the best of our knowledge, no research \eapcomment{has been able to } handle the full (both its structural and parametric part) reconstruction of a general PBN when no prior knowledge is given. On the other hand, in our experiments our method can reconstruct the $2^{1000}$ Markov Chain generated by a $1000$-node network within hours for the case with the smallest amount of training data and after roughly 2.5 days for the case with the largest amount of data. 
\subsection{Experimental Setup}
In subsequent sections, we evaluate the learning and inference methodologies for a 10-\iapcomment{node}, 100-\iapcomment{node} and 1000-\iapcomment{node} model. The true PBNs were generated in the following manner. We used the "generateRandomNKNetwork" function provided by the BoolNet R-tool  \cite{doi:10.1093/bioinformatics/btq124} to randomly generate two deterministic Boolean Networks in a Disjunctive Normal Form (DNF) representation. The deterministic BNs have homogeneous topology (the number of the neighboring nodes is independent and follows a Poisson distribution with mean 4) for each case of network size, and uniform linkage (the edges between the nodes are drawn uniformly at random). The behavior of all the nodes in the resulting network is governed by two Boolean rules, which are dictated by the constituent Boolean networks. Afterwards, we randomly merged the BNs into a PBN by assigning a selection probability to each rule. The selection probability among the rules of the stochastic nodes follows a uniform distribution in [0,1].
\subsection{Recovery of Time Homogeneous Discrete Markov Chain Dynamics}

In order to obtain the time series required for the training (model learning) and testing (accuracy evaluation) of the models, we simulated the true models for a different number of initial conditions, draws for each initial state and time steps for each model size case. Table \ref{tab:samplecomplexity} summarizes all the cases of training sets that were considered.  The final model was learned by applying 5-fold cross-validation \iapcomment{\cite{mitchell1997machine}}, where each fold contains 1/5 of the total number of the training time series. The 10-node, 100-node, \iapcomment{ and} 1000-node models were evaluated on $R=10^4$ different initial conditions. For each one of them $M=400$ stochastic simulations were performed. Note that for the case of the $10$-node network, this number of initial conditions implies exhaustive testing (>$2^{10}$). In all cases, the testing time series contain 1000 time points. The initial conditions of the generated (both training and testing) time series were randomly drawn.

In order to assess the similarity between the true and the learned model, we considered the absolute difference error between the parameters $\mathscr{P}_T^{(i)}\big(\boldsymbol{\lambda}_r,k\big)$ of the true  and the learned model $\mathscr{\hat{P}}_T^{(i)}\big(\boldsymbol{\lambda}_r,k\big)$, where $\boldsymbol{\lambda}_r$ is the vector with the $r$-th randomly drawn initial conditions and $\mathscr{P}_T^{(i)}\big(\boldsymbol{\lambda}_r,k\big)$ the probability that the state of the node $i$ will be $True$ after $k$ steps when the system is initially at the state $\boldsymbol{\lambda}_r$:
\begin{equation}
\mathscr{P}_T^{(i)}\big(\boldsymbol{\lambda}_r,k\big)\triangleq Prob\left(\boldsymbol{\mu}^{(i)}=True,  \boldsymbol{\mu}=\boldsymbol{\Psi}^k(\boldsymbol{\lambda}_r)\right),
\end{equation}
\begin{equation}
\delta^{(i)}_r(k)=\left|\mathscr{P}_T^{(i)}\big(\boldsymbol{\lambda}_r,k\big)-\mathscr{\hat{P}}_T^{(i)}\big(\boldsymbol{\lambda}_r,k\big)\right|.
\end{equation}
Subsequently, the absolute difference was averaged across all the nodes and all the different initial conditions:
\begin{equation}
\bar{\delta}_r(k)=\frac{\sum\limits_{i=1}^{N}\delta^{(i)}_r(k)}{N}, \text{  } \bar{\delta}(k)=\frac{\sum\limits_{r=1}^{R}\bar{\delta}_r(k)}{R}.
\end{equation}
We also considered the standard deviation of the absolute error across the different initial conditions in order to evaluate the dependence of the performance of the model on the initial conditions of the system:
\begin{equation}
\sigma(k)=\sqrt{\frac{\sum\limits_{r=1}^{R}\left(\bar{\delta}_r(k)-\bar{\delta}(k)\right)^2}{R}}.
\end{equation}
We are also reporting the performance of individual nodes by considering the standard deviation of the absolute error given the initial condition and then by averaging the standard deviations across all the initial conditions:
\begin{equation}
\sigma_r(k)=\sqrt{\frac{\sum\limits_{i=1}^{N}\left(\delta^{(i)}_r(k)-\bar{\delta}_r(k)\right)^2}{N}},\text{ }\bar{\sigma}(k)=\frac{\sum\limits_{r=1}^{R}\sigma_r(k)}{R}.
\end{equation}
The parameters $\mathscr{P}_T^{(i)}\big(\boldsymbol{\lambda}_r,k\big)$ and $\mathscr{\hat{P}}_T^{(i)}\big(\boldsymbol{\lambda}_r,k\big)$ were estimated as follows:
\begin{equation}
\label{eq:p_t}
\mathscr{P}_T^{(i)}\big(\boldsymbol{\lambda}_r,k\big)=\frac{\sum\limits_{j=1}^{M}I\big(s^j_k(x_i|\boldsymbol{\lambda}_r)=True\big)}{M},
\end{equation}
\begin{equation}
\label{p_i_k_lambda}
\mathscr{\hat{P}}_T^{(i)}\big(\boldsymbol{\lambda}_r,k\big)=\frac{\sum\limits_{j=1}^{M}I\big(\hat{s}^j_k(x_i|\boldsymbol{\lambda}_r)=True\big)}{M},
\end{equation}
where $s^j_k(x_i|\boldsymbol{\lambda}_r)$ and $\hat{s}^j_k(x_i|\boldsymbol{\lambda}_r)$ are the states of node $i$ at time $k$ for $\boldsymbol{\lambda}_r$ initial system state for the $j$-th stochastic run of the true and learned model, respectively. Note that even for the true model it is computationally intractable to obtain a closed-form value of $\mathscr{P}_T^{(i)}\big(\boldsymbol{\lambda}_r,k\big)$ for large $k$.
\begin{figure*}	
	\centering
     \begin{subfigure}[t]{2.2in}
		\centering
		\includegraphics[width=2.2in, height=2in]{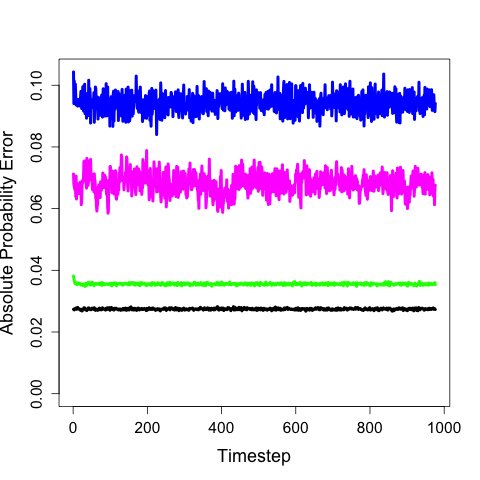}
		\caption{$\bar{\delta}(k)$}\label{fig:1a}		
	\end{subfigure} 
	\begin{subfigure}[t]{2.2in}
		\centering
		\includegraphics[width=2.2in, height=2in]{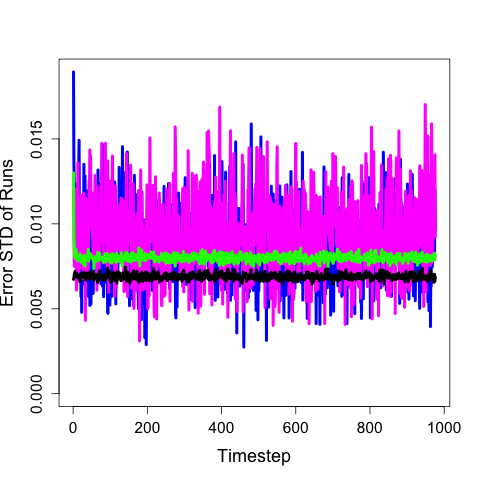}
	\caption{$\sigma(k)$}\label{fig:1b}
	\end{subfigure}
    \begin{subfigure}[t]{2.2in}
		\centering
		\includegraphics[width=2.2in, height=2in]{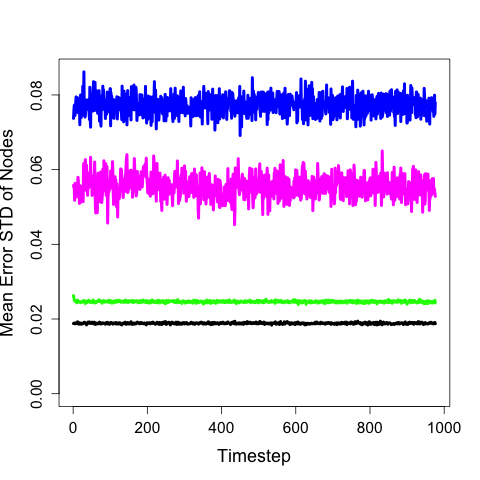} 
		\caption{$\bar{\sigma}(k)$}\label{fig:1c}
	\end{subfigure}
	\caption{Dynamics Prediction of a $2^{10}$-state \iapcomment{discrete-time} homogeneous Markov Chain.}
    \label{fig:10node_learning}
\end{figure*}
\setlength{\belowcaptionskip}{-4pt}
\vspace{-1pt} 
\begin{figure*}	
	\centering
    \begin{subfigure}[t]{2.2in}
		\centering
		\includegraphics[width=2.2in, height=2in]{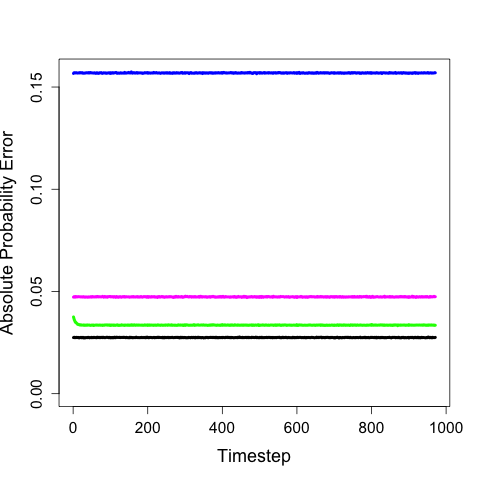}
		\caption{$\bar{\delta}(k)$}\label{fig:1a}		
	\end{subfigure} 
	\begin{subfigure}[t]{2.2in}
		\centering
		\includegraphics[width=2.2in, height=2in]{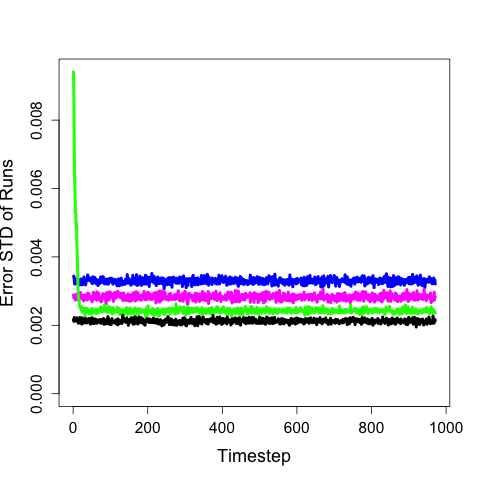}
	\caption{$\sigma(k)$}\label{fig:1b}
	\end{subfigure}
    \begin{subfigure}[t]{2.2in}
		\centering
		\includegraphics[width=2.2in, height=2in]{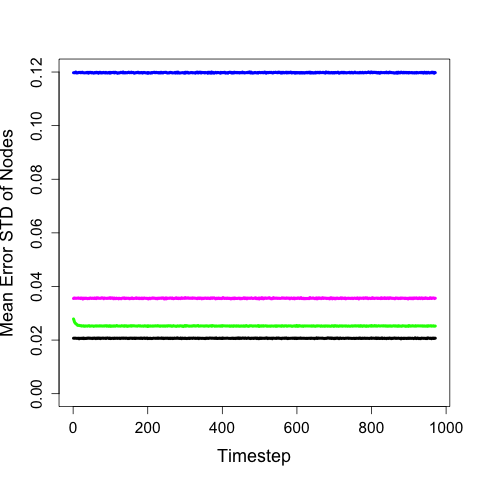} 
		\caption{$\bar{\sigma}(k)$}\label{fig:1c}
	\end{subfigure}
	\caption{Dynamics Prediction of a $2^{100}$-state \iapcomment{discrete-time} homogeneous  Markov Chain.}
       \label{fig:100node_learning}
\end{figure*}
\vspace{-1pt}
\setlength{\belowcaptionskip}{-4pt}
\begin{figure*}	
%
%
	\centering
	\begin{subfigure}[t]{2.2in}
		\centering
		\includegraphics[width=2.2in, height=2in]{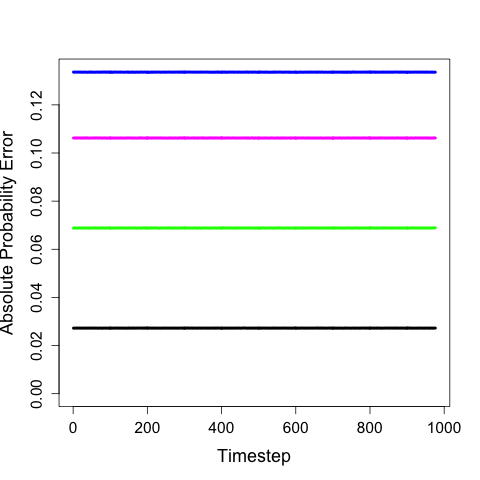}
		\caption{$\bar{\delta}(k)$}\label{fig:1a}		
	\end{subfigure} 
	\begin{subfigure}[t]{2.2in}
		\centering
		\includegraphics[width=2.2in, height=2in]{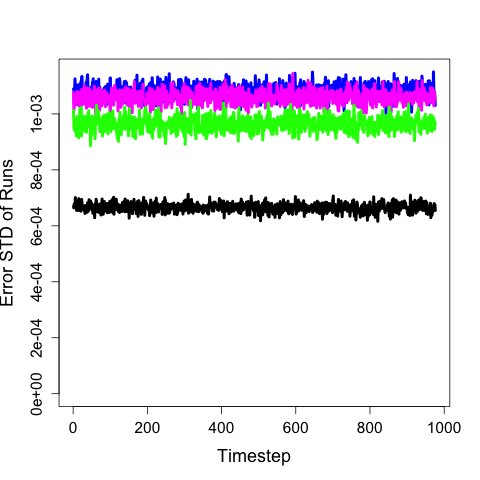}
	\caption{$\sigma(k)$}\label{fig:1b}
	\end{subfigure}
    \begin{subfigure}[t]{2.2in}
		\centering
		\includegraphics[width=2.2in, height=2in]{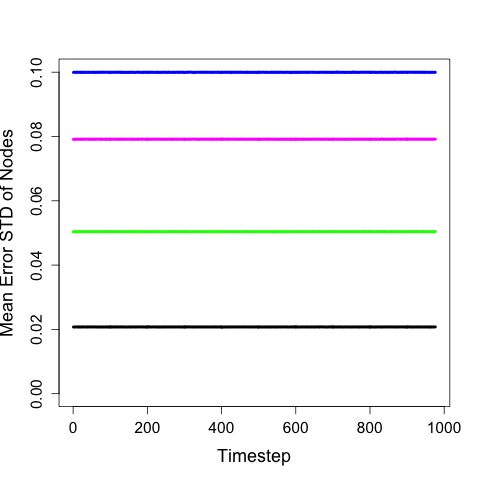} 
		\caption{$\bar{\sigma}(k)$}\label{fig:1c}
	\end{subfigure}
	\caption{Dynamics Prediction of a $2^{1000}$-state \iapcomment{discrete-time} homogeneous Markov Chain.}
  \label{fig:1000node_learning}
\end{figure*}

\begin{table*}
\centering
\caption{Sample Complexity.}
\begin{threeparttable}
  \begin{tabular}{|c|c|c|}
   \hline
    10-node & 100-node &1000-node\\
     \hline
     \begin{tikzpicture}
    	\begin{customlegend}[legend entries={20/5/4, 20/5/8, 20/5/16, real model}]
    	\addlegendimage{blue}
    	\addlegendimage{magenta}
    	\addlegendimage{green}
    	\addlegendimage{black}
    	\end{customlegend}
	\end{tikzpicture} 
	&
	\begin{tikzpicture}
    	\begin{customlegend}[legend entries={100/10/2, 200/10/4, 200/10/8, real model}]
    	\addlegendimage{blue}
    	\addlegendimage{magenta}
    	\addlegendimage{green}
    	\addlegendimage{black}
    	\end{customlegend}
	\end{tikzpicture} 
	&
	\begin{tikzpicture}
    	\begin{customlegend}[legend entries={125/25/2, 1000/50/2, 8000/400/2, real model}]
    	\addlegendimage{blue}
    	\addlegendimage{magenta}
   		\addlegendimage{green}
    	\addlegendimage{black}
    	\end{customlegend}
	\end{tikzpicture} \\
   \hline
  \end{tabular}
   \begin{tablenotes}\footnotesize
\item Each case refers to the [number of timeseries/ number of different initial conditions/ number of timepoints]  in the training set. 
\end{tablenotes}

   \end{threeparttable}

   \label{tab:samplecomplexity}
        
\end{table*}
The quantity $N \times \bar{\delta}(k)$ can be interpreted as the average number of wrongly predicted node states at time $k$, regardless the initial state. Alternatively, other difference metrics such as the Kullback Leibler divergence and the Hellinger distance have been investigated. Here, we only report the absolute difference given its direct and intuitive interpretation. Also note that only the initial state is given to the network (and that the predicted states are used for the prediction of the next state). Figures \ref{fig:10node_learning}, \ref{fig:100node_learning}, \ref{fig:1000node_learning} depict the predictive capacity achieved by the learned models. The reader may refer to Table \ref{tab:samplecomplexity}, for the description of the coloring schemes in these plots \iapcomment{ which have to do with the various sizes of the training sets generated for the learning}. The dark line refers to the unavoidable error, which occurs during the simulation of the true model, due  to the finite $M$ for the estimation of $\mathscr{P}_T^{(i)}\big(\boldsymbol{\lambda}_r,k\big)$ in Equation \ref{eq:p_t}. These estimated parameters were obtained from a second \iapcomment{round} of stochastic simulation\iapcomment{s} of the real model.

As it can be concluded, the SCNF has managed to capture, almost perfectly, the dynamical behavior of the true PBN for $N=10$ and $N=100$ (the black and green lines in Figures \ref{fig:10node_learning}.a, \ref{fig:100node_learning}.a almost overlap) with similarly low standard deviation of the misprediction rate for the randomized initial conditions used for the testing (black and green lines in Figures \ref{fig:10node_learning}.b, \ref{fig:100node_learning}.b). The standard deviation of the predictability of individual nodes almost fits the performance of the true model (Figures \ref{fig:10node_learning}.c, \ref{fig:100node_learning}.c). Especially in the case of the 10-node network \eapcomment{which} is learned with only 4 or 8 time steps (blue and pink line), the resulting formula for many nodes is simply a deterministic CNF formula. This fact is responsible for the high variance in the performance of individual nodes (\ref{fig:10node_learning}.c) and in the performance of the model for different initial conditions (\ref{fig:10node_learning}.a, \ref{fig:10node_learning}.b). 

In the 1000-node case, the best average number of mispredicted nodes in all time points is roughly $0.07 \times 1000$ (green line in Figure \ref{fig:1000node_learning}.a) compared to the performance of the true model, which is $0.03 \times 1000$ (black line in Figure \ref{fig:1000node_learning}.a). However, as it is corroborated by the blue and pink lines in Figures \ref{fig:1000node_learning}.a, \ref{fig:1000node_learning}.b, \ref{fig:1000node_learning}.c, a network learned with only 125 single-step temporal sequences can still exhibit high accuracy. 

We also proceeded to the reconstruction of a \iapcomment{$10^4$-}node network which achieved error roughly $0.9$ for $R=100$. However, due to limited computational resources we were unable to test its performance for larger $R$, therefore we omitted these results.
\subsection{Transition Probabilities Estimation}

We now present the results on the networks learned in order to evaluate the computational cost and convergence of the sampling scheme described above \iapcomment{(Equation \ref{p_i_k_lambda})}. In all cases of network sizes $N=10, 100, 1000$, we inferred 2-step and 100-step transition probabilities (parameters) $\mathscr{P}^{(i)}_T\big(\boldsymbol{\lambda}_r,k\big)$, for $\boldsymbol{\lambda}_r \in \{False, True\}^N, i=1,2, \dots, N$, $k=2, 100$. Note that this implies also the inference of all the $2^N$, $k$-step transition probabilities of moving from the state $\boldsymbol{\lambda}_r$ to the state 
$\boldsymbol{\mu}=\{False,True\}^N$\iapcomment{, i.e}
$\mathscr{P}^{(i)}\big(\boldsymbol{\mu}|\boldsymbol{\lambda}_r,k\big)$, where:
\begin{equation}
\mathscr{P}\big(\boldsymbol{\mu}|\boldsymbol{\lambda}_r,k\big)\triangleq Prob\left(\boldsymbol{\Psi}^k(\boldsymbol{\lambda}_r)=\boldsymbol{\mu}\right),
\end{equation}
since it holds that:
\begin{equation}
\label{eq:hat_p_system}
\mathscr{P}\big(\boldsymbol{\mu}|\boldsymbol{\lambda}_r,k\big) =\mathlarger{\prod}\limits_{i=1}^{N}\mathscr{P}^{(i)}_T\big(\boldsymbol{\lambda}_r,k\big)^{\mu^{(i)}}\big(1-\mathscr{P}^{(i)}_T\big(\boldsymbol{\lambda}_r,k\big)\big)^{1-\mu^{(i)}}, 
\end{equation}
where for ease of notation $\mu^{(i)}$ is the numerical equivalent $1/0$ of its \iapcomment{boolean} value $True/False$. \iapcomment{More specifically, the convergence of the transition probabilities at a node level in Equation \ref{p_i_k_lambda} is guaranteed by the Central Limit Theorem according; that is, the estimation will converge to the normal distribution $\mathscr{N}\big(\mathscr{P}_T^{(i)}\big(\boldsymbol{\lambda}_r,k\big), \mathscr{P}_T^{(i)}\big(\boldsymbol{\lambda}_r,k\big)(1-\mathscr{P}_T^{(i)}\big(\boldsymbol{\lambda}_r,k\big)\big)$ for large $M$. Eventually, the convergence of the transition probability at the system level in Equation \ref{eq:hat_p_system} stems from the Multivariate Delta Method \iapcomment{\cite{wasserman2013all}.} }

\begin{figure}
	\centering
	\begin{subfigure}{2.5in}
		\centering	\includegraphics[width=2.5in,height=1.6in]{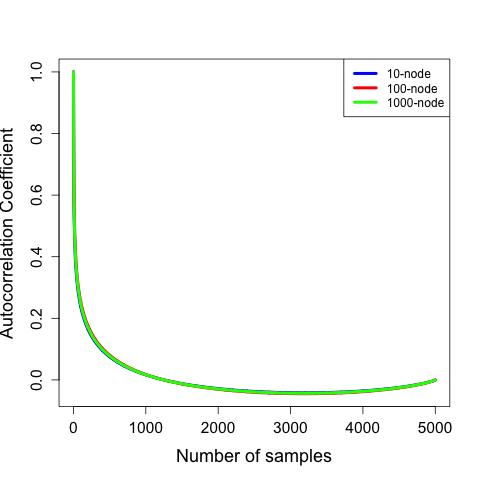}
		\caption{2-step transitions}
 \label{fig:2stepinferconver}         \hspace{1.05\baselineskip}
	\end{subfigure} 
	\begin{subfigure}{2.5in}
		\centering	\includegraphics[width=2.5in,height=1.6in]{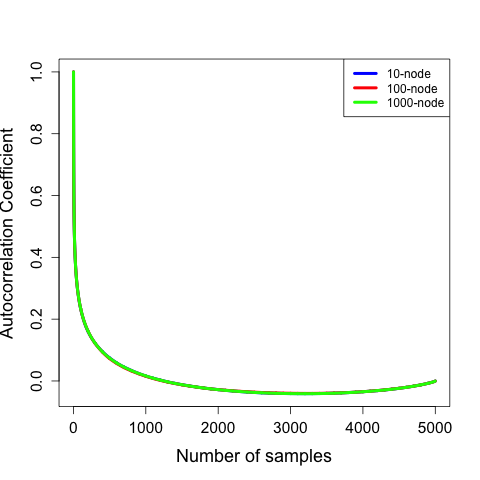}
		\caption{100-step transitions}
        \label{fig:100stepinferconver}
	\end{subfigure}
    \hspace{-1.05\baselineskip}
	\caption{Inference Convergence for 2-step and 100-step transitions.}
    \label{fig:inferenceconvergence}
\end{figure}
\begin{figure}
	\centering
	\begin{subfigure}[t]{2.5in}
		\centering
		\includegraphics[width=2.5in,height=1.6in]{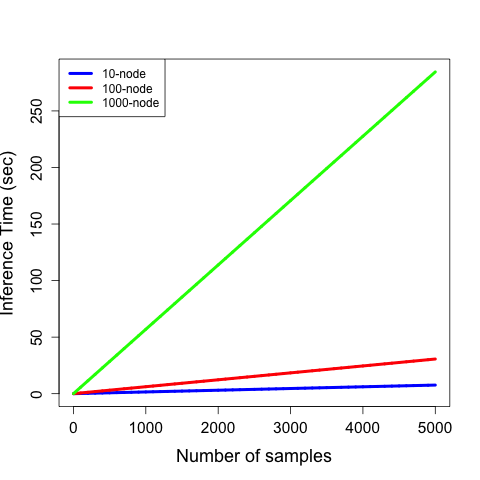}
		\caption{2-step transitions}
        \label{fig:2stepinfera}		
      \hspace{1.05\baselineskip}
	\end{subfigure} 
	\begin{subfigure}[t]{2.5in}
		\centering
		\includegraphics[width=2.5in,height=1.6in]{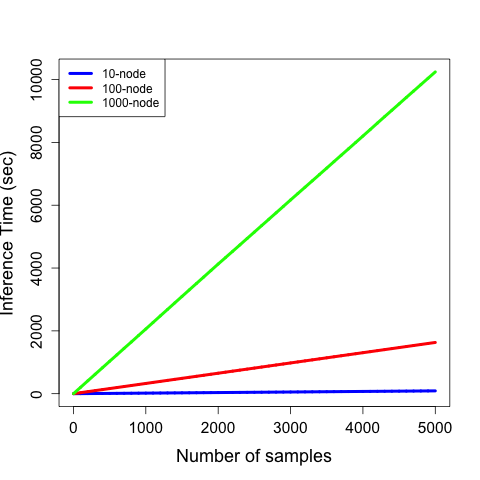}
		\caption{100-step transitions}
        \label{fig:2stepinferb}
	\end{subfigure}
	\caption{Inference Run
    time for 2-step and 100-step transitions.}
    \label{fig:inferenceruntime}
\end{figure}

The initial system states $\boldsymbol{\lambda}_r$ were uniformly selected out of the $2^N$ possible logical values. The results reported are averaged across  $R=10^6$ and $R=10^4$ iterations (different cases of \iapcomment{$\boldsymbol{\lambda}_r$} initial conditions) of stochastic simulations for the 2-step and 100-step transitions, respectively. At this point, we should mention that we had to reduce the number of stochastic simulations for the $100$-step parameters due to limited computational resources. However, the convergence of the parameters does not depend on the number of transition time steps \iapcomment{$k$} considered. $M=5000$ samples were drawn so as to approximate the transition probabilities in each stochastic simulation. We tested the convergence of parameters for individual nodes $\mathscr{P}^{(i)}_T\big(\boldsymbol{\lambda}_r,k\big)$. As a convergence diagnostic metric, we use the mean autocorrelation coefficient: 
\begin{equation}
\label{eq:acf}
\rho(lag)=\frac{\sum\limits_{r=1}^{R}\rho_r({lag})}{R},
\end{equation}
where:
\begin{equation}
\rho_r(lag)=\frac{\sum\limits_{i=1}^{N}\rho_r^{(i)}({lag})}{N}
\end{equation}
for $lag=1,2,\dots,M$.
\iapcomment{$\rho_r^{(i)}({lag})$} is the autocorrelation coefficient for \iapcomment{the} parameter \iapcomment{$\mathscr{P}^{(i)}_T\big(\boldsymbol{\lambda}_r,k\big)$} of the individual node $i$, and it is defined as: 
\begin{align}
\label{acf}
& 
\rho_r^{(i)}({lag})=\frac{\mathlarger{\sum}\limits_{m=1}^{M-lag}\big(\hat{p}^{(i)}_r(m)-\bar{\hat{p}}_r^{(i)}\big)\big(\hat{p}^{(i)}_r(m+lag)-\bar{\hat{p}}^{(i)}_r\big)}{\mathlarger{\sum}\limits_{m=1}^{M-lag}\big(\hat{p}^{(i)}_r(m)-\bar{\hat{p}}^{(i)}_r\big)^2},
\nonumber \\
& \hat{p}^{(i)}_r(m) \text{ defined in Equation \ref{p_i_k_lambda} for } M=m, \nonumber \\
& \bar{\hat{p}}^{(i)}_r=\frac{\sum\limits_{m=1}^{M}\hat{p}^{(i)}_r(m)}{M}.
\end{align}

In the above formula $lag=0$, implies $\rho^{(i)}_r(lag)=1$. The number of samples ($lag$) for the inference is sufficiently large, once the estimations $\hat{p}^{(i)}_r(m), \hat{p}^{(i)}_r(m+lag)$ are uncorrelated. Finally, we average $\rho_r(lag)$ across all $R$ iterations (see Equation \ref{eq:acf}). In all cases of network sizes and steps, the autocorrelation coefficient approaches zero after roughly 2000 samples and stabilizes there beyond that (see Figure \ref{fig:inferenceconvergence}). Furthermore, according to Figure \ref{fig:inferenceruntime}, the inference time grows linearly with respect to the number of steps and the number of nodes in the network. At this point we would like to point out \iapcomment{that} we obtained results for $k=1000$ and $R=10^3$ which exhibited the same trend as in Figures \ref{fig:inferenceconvergence}, \ref{fig:inferenceruntime}. However, due to limited computational resources, we could not simulate the model for larger $R$, therefore we omitted these results. 

Finally, it should be \iapcomment{emphasized} that the inference procedure is inherently parallel since the disjunctions are independently activated and the logical value can be directly computed if at least one disjunction is evaluated as $False$. However, the inference time in Figure \ref{fig:inferenceruntime} refers to the sequential execution time so that it can better illustrate the computational demands of the sampling method.
\section{Illustrative Examples}
\begin{table*}
\caption{Derivation of the transition probabilities for each node in the Example 1.}
  \centering
  \resizebox{0.8\textwidth}{!}{
  \renewcommand{\arraystretch}{0.8}
  \begin{tabular}{|c c c||c|c|c||c|c||c|c|c|}
    \hline
    $x_1$ & $x_2$ & $x_3 $ & $x_2 \lor \neg{x}_1 \land x_3$ & $x_1\lor x_2$ & $Prob(x_1'=T)$ & $x_2 \land \neg{x}_1$ & $Prob(x_2'=T)$ & $x_1 \lor x_3$ & $\neg{x}_3$ & $Prob(x_3'=T)$ \\
    \hline
    F & F & F & F & F & 0.0 & F & 0.0 & F & T & 0.2\\
    F & F & T & T & F & 0.6 & F & 0.0 & T & F & 0.8\\
    F & T & F & T & T & 1.0 & T & 1.0 & F & T & 0.2\\
    F & T & T & T & T & 1.0 & T & 1.0 & T & F & 0.8\\
    T & F & F & F & T & 0.4 & F & 0.0 & T & T & 1.0\\
    T & F & T & F & T & 0.4 & F & 0.0 & T & F & 0.8\\
    T & T & F & T & T & 1.0 & F & 0.0 & T & T & 1.0\\
    T & T & T & T & T & 1.0 & F & 0.0 & T & F & 0.8\\
    \hline
  \end{tabular}
  }
  \label{tab:example1}
\end{table*}
\begin{figure}
	\centering
\includegraphics[width=150pt, height=100pt]{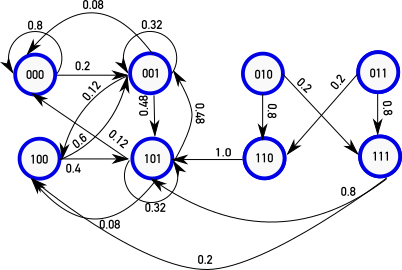}
	\caption{The probability transition diagram of Example 1. }
	\label{fig:example1}
\end{figure}
\subsection{A simple PBN as a SCNFN}
Consider a PBN which is given by the following rules:
\allowdisplaybreaks
\begin{equation}
\allowdisplaybreaks
\begin{aligned}
\label{eq:f_x}
& \mathbf{f}^{(A)} = \begin{cases} f^{(A)}_1=x_2\lor\neg{x}_1\land x_2, \ p^{(A)}_1=0.6 \\
f^{(A)}_2=x_1 \lor x_2, \  p^{(A)}_2=0.4, \end{cases} &\\
& f^{(B)}=x_2\neg{x}_1 \ p_{2_1}=1.0, &\\
& \mathbf{f}^{(C)}=\begin{cases} f^{(C)}_1=x_1\land x_3, \ p^{(C)}_1=0.8 \\
f^{(C)}_2=\neg{x}_3, \ p^{(C)}_2=0.2. \end{cases}
\end{aligned}
\end{equation}
The computation of the transition probabilities of each node $i$ is demonstrated in Table \ref{tab:example1}. Subsequently, the probability transition diagram can be computed, as illustrated in Figure \ref{fig:example1}.
The rules of the nodes $x_1, x_2, x_3$,  written in the form of SCNF and according to Table \ref{tab:example1} are:
\allowdisplaybreaks
\begin{align}
& \tilde{\Psi}^{(1)} = \nonumber \\ 
& \qquad \ \ (x_1\lor x_2 \lor x_3)\land (x_1 \lor x_2 \lor \neg{x}_3 \lor \alpha^{(1)}_1) \nonumber \\ 
& \qquad {\land}\:(\neg{x}_1\lor x_2 \lor x_3 \lor \alpha^{(1)}_2) \nonumber \\ 
& \qquad{\land}\:(\neg{x}_1 \lor x_2 \lor \neg{x}_3 \lor \alpha^{(1)}_3), \nonumber \\ 
& q^{(1)}_1=0.4, q^{(1)}_2=0.6, q^{(1)}_3=0.6, 
\label{eq:f_x1} \\ 
& \tilde{\Psi}^{(2)} = \nonumber \\ 
& \qquad \ \ (x_1\lor x_2 \lor x_3)\land (x_1\lor x_2\iapcomment{\lor}\neg{x}_3) \nonumber\\ 
& \qquad {\land}\:(\neg{x}_1\lor x_2\lor x_3)\land (\neg{x}_1\lor x_2\lor \neg{x}_3) \nonumber\\ 
& \qquad {\land}\:(\neg{x}_1\lor \neg{x}_2\lor x_3)\land(\neg{x}_1\lor \neg{x}_2\lor \neg{x}_3), 
\label{eq:f_x2} \\
& \tilde{\Psi}^{(3)} = \nonumber \\ 
& \qquad \ \ (x_1\lor x_2\lor x_3 \lor \alpha^{(3)}_1)  \nonumber \\ 
& \qquad {\land}\:(x_1\lor x_2\lor \neg{x}_3 \lor \alpha^{(3)}_2) \nonumber \\
& \qquad {\land}\:(x_1\lor \neg{x}_2\lor x_3 \lor \alpha^{(3)}_3) \nonumber \\
& \qquad {\land}\:(x_1\lor \neg{x}_2\lor \neg{x}_3 \lor \alpha^{(3)}_4) \nonumber \\
& \qquad {\land}\:(\neg{x}_1\lor x_2\lor x_3 \lor \alpha^{(3)}_5 ) \nonumber \\
& \qquad {\land}\:(\neg{x}_1\lor \neg{x}_2\lor \neg{x}_3 \lor \alpha^{(3)}_6),\nonumber \\ 
& q^{(3)}_1=0.8,  q^{(3)}_2=0.2, q^{(3)}_3=0.8, \nonumber \\
& q^{(3)}_4=0.2,  q^{(3)}_5=0.2, q^{(3)}_6=0.2.\label{eq:f_x3}
\end{align}

We will now describe the conversion for the node $x_1$. The other two SCNF rules can be derived in a similar fashion. The first disjunction $x_1 \lor x_2 \lor x_3$ is deterministic and always evaluated since for the state $\iapcomment{\boldsymbol{\lambda}}=[F, F, F]$ both of the rules yield $False$. The disjunctions $x_1 \lor \neg{x}_2\lor x_3$, $x_1 \lor \neg{x}_2 \lor \neg{x}_3$,   $\neg{x}_1 \lor \neg{x}_2\lor x_3$, and $\neg{x}_1 \lor \neg{x}_2 \lor \neg{x}_3$ are omitted since both of the rules are evaluated as $True$ for the states $[0,1,0], [0,1,1], [1,1,0]$, and $[1,1,1]$ accordingly. The parameter of the disjunction $x_1\lor x_2 \lor \neg{x}_3$ is 0.4 since the next state of $x_1$ when the current state is $[0,0,1]$ is $True$ with probability $0.6$ (see the second row of Table \ref{tab:example1}). Observe that the rest of the stochastic disjunctions will always be $True$ for this state, regardless of the outcome of their associated Bernoulli variables. \iapcomment{Afterwards}, the next state of $x_1$ should be $False$ with probability $0.6$ and for current state $[1,0,0]$. This negative transition is satisfied by the disjunction $\neg{x}_1 \lor x_2 \lor x_3$. Finally, the disjunction $\neg{x}_1 \lor x_2 \lor \neg{x}_3$, carries the information that when the current state is $[1,0,1]$, $x_1$ goes to $False$ with probability 0.4.
\begin{table*}[ht]
\caption{Example 2. : Steps of the learning algorithm.}
\centering
  \resizebox{0.8\textwidth}{!}{
  \renewcommand{\arraystretch}{0.2}
\begin{tabular}
{| c|| c| c| c|}
\hline
 Step & $\Psi$ & $H_F$ & $H_T$ \\\hline
 0 & $\emptyset$ & $\{\boldsymbol{\lambda}_1,\boldsymbol{\lambda}_3,\boldsymbol{\lambda}_5,\boldsymbol{\lambda}_6\}$ & $\{\boldsymbol{\lambda}_0, \boldsymbol{\lambda}_2, \boldsymbol{\lambda}_4,\boldsymbol{\lambda}_7, \boldsymbol{\lambda}_8\}$ \\
 1 & $(\neg{B}$ & $\{\boldsymbol{\lambda}_1,\boldsymbol{\lambda}_3,\boldsymbol{\lambda}_5,\boldsymbol{\lambda}_6\}$ &  $\{\boldsymbol{\lambda}_2, \boldsymbol{\lambda}_4\}$ \\
 2 & $(\neg{B} \lor\neg{A})$ & $\{\boldsymbol{\lambda}_1,\boldsymbol{\lambda}_6\}$ & $\{\boldsymbol{\lambda}_0, \boldsymbol{\lambda}_2, \boldsymbol{\lambda}_4,\boldsymbol{\lambda}_7, \boldsymbol{\lambda}_8\}$ \\
 3 & $(\neg{B} \lor \neg{A})\land (E$ & $\{\boldsymbol{\lambda}_1,\boldsymbol{\lambda}_6\}$ & $\{\boldsymbol{\lambda}_8\}$ \\
 4 & $(\neg{B} \lor \neg{A})\land(E\lor \neg{B})$ & $\{\boldsymbol{\lambda}_0\}$ & $\{ \boldsymbol{\lambda}_2, \boldsymbol{\lambda}_4,\boldsymbol{\lambda}_7, \boldsymbol{\lambda}_8\}$ \\
 5 & $(\neg{B} \lor \neg{A})\land(E\lor\neg{B})\land(\neg{A}$ & $\{\boldsymbol{\lambda}_0\}$ & $\{\boldsymbol{\lambda}_8\}$ \\
 6 & $(\neg{B} \lor \neg{A})\land(E\lor\neg{B})\land (\neg{A}\lor\neg{E} \lor \alpha^{(A)}_1)$ & $\emptyset$ &$\emptyset$  \\ 
 7 & $(\neg{B} \lor \neg{A})\land(E\lor\neg{B})\land(\neg{A}\lor\neg{E} \lor \alpha^{(A)}_1), p^{(A)}_1=0.67$ & $\emptyset$ &$\emptyset$  \\ 
\hline
  \end{tabular}
  }
  \label{tab:example2learning}
\end{table*}
\subsection{Learning a simple SCNF formula from a single time series}
\begin{table}
\caption{Example 2. :  Sets $\mathsf{L}, \color{yellow} \mathsf{S}^{(A)}_C, \color{red} \mathsf{S}^{(A)}_F, \color{green} \mathsf{S}^{(A)}_T$.}
\centering
\resizebox{\columnwidth}{!}{
 \renewcommand{\arraystretch}{0.3}
\begin{tabular}
{c c c c c c c c c c c c}
\hline
  $\mathbf{S}_t$ & & $A$ & $B$ & $C $ & $D$ & $E$ & $F$ & $G$ & $H$ & $I$ & $J$ \\
\hline
\rowcolor{yellow}
$\mathbf{S}_0$ &$\boldsymbol{\lambda}_0$ & T & F & T & T & T & F & T & T & T & F \\
\rowcolor{red}
$\mathbf{S}_1$ &$\boldsymbol{\lambda}_1$ & F & T & T & T & F & F & F &F & F & T \\
\rowcolor{green}
$\mathbf{S}_2$ &$\boldsymbol{\lambda}_2$& F & T & F & T & T & F & F & F & F & T \\
\rowcolor{red}
$\mathbf{S}_3$ &$\boldsymbol{\lambda}_3$& T & T & F & F & T & T & F & F & F & F \\
\rowcolor{green}
$\mathbf{S}_4$ &$\boldsymbol{\lambda}_4$& F & T & F & F & T & T & F & F & F & F \\
\rowcolor{red}
$\mathbf{S}_5$ &$\boldsymbol{\lambda}_5$& T & T & F & F & T & T & F & F & F & T \\
\rowcolor{red}
$\mathbf{S}_6$ &$\boldsymbol{\lambda}_6$& F & T & F & F & F & T & F & F & F & F \\
\rowcolor{green}
$\mathbf{S}_7$ &$\boldsymbol{\lambda}_7$& F & F & T & T & T & F & T & T & T & T \\
\rowcolor{yellow}
$\mathbf{S}_8$ &$\boldsymbol{\lambda}_0$&T & F & T & T & T & F & T & T & T & F \\
\rowcolor{green}
$\mathbf{S}_9$ &$\boldsymbol{\lambda}_8$& T & F & T & T & F & F & T & T & T & F \\
\rowcolor{yellow}
$\mathbf{S}_{10}$ &$\boldsymbol{\lambda}_0$ & T & F & T & T & T & F & T & T & T & F \\
 $\mathbf{S}_{11}$ &$\boldsymbol{\lambda}_9$& F & F & T & T & F & F & T & T & T & T \\
    \hline
  \end{tabular}
 }
  \label{tab:example2timeseries}
\end{table}
Consider a set $\mathsf{D}$ which consists of a single time series with $10$ transitions provided in Table \ref{tab:example2timeseries}.  The \iapcomment{list} $\mathsf{L}^{(A)}$ for the node A, and the sets $\mathsf{S}^{(A)}_F$,  $\mathsf{S}^{(A)}_T$,  $\mathsf{S}^{(A)}_C$ are:
\allowdisplaybreaks
\begin{align}
& \mathsf{L}= \big((\mathbf{S}_0,\mathbf{S}_1), (\mathbf{S}_1,\mathbf{S}_2), (\mathbf{S}_2,\mathbf{S}_3), \nonumber \\  
 & \qquad \ (\mathbf{S}_3,\mathbf{S}_4), (\mathbf{S}_4,\mathbf{S}_5), (\mathbf{S}_5,\mathbf{S}_6),  \nonumber \\ 
& \qquad \ (\mathbf{S}_6,\mathbf{S}_7), (\mathbf{S}_7,\mathbf{S}_8), (\mathbf{S}_8,\mathbf{S}_9), \nonumber \\ 
& \qquad \ (\mathbf{S}_9, \mathbf{S}_{10}), (\mathbf{S}_{10},\mathbf{S}_{11})\big), & \\ 
& \mathsf{L}^{(A)}= \big((\mathbf{S}_0,F), (\mathbf{S}_1,F), (\mathbf{S}_2,T), \nonumber \\  
 & \qquad \ \ \ \ (\mathbf{S}_3,F), (\mathbf{S}_4,T), (\mathbf{S}_5,F),  \nonumber \\ 
& \qquad \ \ \ \ (\mathbf{S}_6,F), (\mathbf{S}_7,T), (\mathbf{S}_8,T), \nonumber \\ 
& \qquad \ \ \ \ (\mathbf{S}_9, T), (\mathbf{S}_{10},F)\big), & \\ 
& \mathsf{S}^{(A)}_F= \{ 
 \boldsymbol{\lambda}_1,\boldsymbol{\lambda}_3,\boldsymbol{\lambda}_5,\boldsymbol{\lambda}_6 \nonumber
\}, 
\mathsf{S}^{(A)}_T= \{
\boldsymbol{\lambda}_2, \boldsymbol{\lambda}_4, \boldsymbol{\lambda}_7, \boldsymbol{\lambda}_8
\}, \nonumber \\
& \mathsf{S}^{(A)}_C= \{\boldsymbol{\lambda}_0\}.
\end{align}
Note that, the transition $\boldsymbol{\lambda}_0$ is a conflict, since at time 1 and 11, the node A has state $False$ when the system state is $\boldsymbol{\lambda}_0$, while at time 9, the node A has state $True$. Table \ref{tab:example2scores} provides the scores of the literals for each step of the learning process. We will provide the details for the computation of the score of $\neg{B}$ in the first step. (1) $score^{+}(\neg{B})=3$, for the states $\boldsymbol{\lambda}_0,\boldsymbol{\lambda}_7,\iapcomment{\boldsymbol{\lambda}_8}$, (2) $|H_T|=\big|\mathsf{S}^{(A)}_T\big|+\big|\mathsf{S}^{(A)}_C\big|=4+1=5$, (3) $score^{-}(\neg{B})=0$, (4) $|H_F|=4 $, (5) $score(\neg{B})=\frac{3}{5}-\frac{0}{4}$. The step-by-step construction of the SCNF can be found in Table \ref{tab:example2learning}. \iapcomment{Note that $\boldsymbol{\lambda}_0$ is initially in $H_T$ and once the reconstruction of the stochastic part starts at Step 4, it is moved to $H_F$. }After the inclusion of the first literal $\neg{B}$, the states $\boldsymbol{\lambda_0}, \boldsymbol{\lambda_7}, \boldsymbol{\lambda_8}$ are removed from $H_T$ since the disjunction will be evaluated as $True$ due to the presence of $\neg{B}$. On the other hand $H_F$ remains the same since $\neg{B}$ does not become $True$ for any transition contained in it. After the inclusion of the literal $\neg{A}$, all the transitions in $H_T$ are evaluated as $True$ and the algorithm proceeds to the formation of the next disjunction. The disjunction $(\neg{B} \lor \neg{A})$ yields $False$ for the states $\boldsymbol{\lambda_3}, \boldsymbol{\lambda_5}$, therefore the next disjunction should give $False$ only for the states $\boldsymbol{\lambda_1}, \boldsymbol{\lambda_6}$. The learning of the deterministic logic part finishes when $H_F$ becomes empty. Steps 5 and 6 pertain to the learning of the stochastic logic part, which proceeds in the same way once the new sets $H_T$ and $H_F$ are formed. Step 7 in Table \ref{tab:example2learning}, refers to the learning of the parametric part of the SCNF model. The relevant quantities are:\\
$
\mathscr{N}^{(A)}_F(\boldsymbol{\lambda}_0)=2, 
\mathscr{N}^{(A)}_T(\boldsymbol{\lambda}_0)=1, \\
\tilde{\Psi}^{(A)}_F(\boldsymbol{\lambda}_0)=\big\{\neg{A} \lor \neg{E} \lor \alpha^{(A)}_1\big\}, \\
\mathscr{P}^{(A)}_F\big(\boldsymbol{\lambda}_0\big)=p^{(A)}_1, \varepsilon_A(\boldsymbol{\lambda}_0)=\mathscr{P}^{(A)}_F\big(\boldsymbol{\lambda}_0\big)-p^{(A)}_1=0, \nonumber\\
\ell_A(\boldsymbol{\lambda}_0)=2*log\big(p^{(A)}_1\big)+1*log\big(1-p^{(A)}_1\big), p^{(A)}_1=0.67.
$
\begin{table}
\centering
  \resizebox{0.9\columnwidth}{!}{
  \begin{threeparttable}
\caption{Example 2. : Scores of the literals.}
\label{tab:example2scores}
\renewcommand{\arraystretch}{0.3}
\begin{tabular}
{| c|| c| c| c| c| c| c|}
\hline
   & 1 & 2  & 3 & 4 & 5 & 6  \\
\hline \hline
  $\neg{A}$ & \ 0.10 & $\mathbf{\ 0.5}$ & -0.40 & * & $\mathbf{\ 0.75}$ & *  \\
\hline
  $\neg{B}$ & $\mathbf{\ 0.60}$ & * & \ 0.60  & $\mathbf{\ 1.00}$ & -0.50 & \ 0.00\\
\hline
  $\neg{C}$ & -0.35 & \ 0.25 &  -0.10  & * & \ 0.50 & *\\
\hline
  $\neg{D}$ & -0.55 & -0.25 &  -0.30  & * & \ 0.25 & *\\
\hline
  $\neg{E}$ & -0.30 & * &  -0.80  & * & \ 0.25 & $\mathbf{\ 1.00}$ \\
\hline
  $\neg{F}$ & \ 0.55 & \ 0.25 & \ 0.30  & \ 0.50 & -0.25 & \ 0.00\\
\hline
  $\neg{G}$ & -0.60 & 0.00 & -0.60 & *  & \ 0.50 & *\\
\hline
  $\neg{H}$ & -0.60 & 0.00 & -0.60  & * & \ 0.50 & *\\
\hline
  $\neg{I}$ & -0.60 & 0.00 & -0.60  & * & \ 0.50 & *\\
\hline
  $\neg{J}$ & \ 0.10 & 0.00 &  0.10  & \ 0.50 & -0.50 & \ 0.00\\
  \hline
  $A$ & -0.10 & * &  \ 0.40   & \ 1.00 & -0.75 & \ 0.00 \\
\hline
  $B$ & -0.60 & * & -0.60   & * & \ 0.50 & \ 0.00 \\
\hline
  $C$ & \ 0.35 & * &  \ 0.10  & \ 0.50 & -0.50 & \ 0.00\\
\hline
  $D$ & \ 0.55 & \ 0.25 &  \ 0.30  & \ 0.50 & -0.25 & \ 0.00\\
\hline
  $E$ & \ 0.30 & \ 0.50 & $\mathbf{\ 0.80}$ & * & -0.25 & *\\
\hline
  $F$ & -0.55 & -0.25 &  -0.30  & * & \ 0.25 & *\\
\hline
  $G$ & \ 0.60 & * &  \ 0.60 &  \ 1.00 & -0.50 & \ 0.00\\
\hline
  $H$ & \ 0.60 & * &  \ 0.60 & \ 1.00 & -0.50 & \ 0.00 \\
\hline
  $I$ & \ 0.60 & * &  \ 0.60 & \ 1.00 & -0.50 & \ 0.00 \\
\hline
  $J$ & -0.10 & \ 0.00 & -0.10 & *  & \ 0.50 & * \\
    \hline
  \end{tabular}
  \label{tab:example2literals}
     \begin{tablenotes}\footnotesize
\item The score of the literals for each step of learning the SCNF of node A. The * denotes that the score of the literal is not computed either because it (or its negation) is already being used in the current disjunction, or because its positive score is zero (it does not evaluate as True any remaining positive transition.). With bold black we annotate the score of the selected literal.
\end{tablenotes}
   \end{threeparttable}
   }
\end{table}
\section{Conclusion}
We have introduced a novel modeling framework of Probabilistic Boolean Networks\eapcomment{, namely }the Stochastic Conjunctive Normal Form Network. We proved that both PBN and SCNFN can represent the same class of \iapcomment{boolean} relationships between the nodes of a network. The adoption of PBNs by a wide spectrum of scientific domains has stimulated an intense research effort on network identification in recent years. However, the current approaches face one or more of the following limitations:
\begin{enumerate}
\item Prior knowledge of the network structure and/or the node interactions is assumed so that the parametric portion of the PBN; whose contribution is critical if the dynamics generated by the PBN have to be predicted, can be estimated from observed time series data.
\item Only the candidate \iapcomment{boolean} rules which regulate the behavior of a node are learned, while the estimation of the corresponding selection probabilities is ignored.
\item Very strong assumptions are made on the structure of the \iapcomment{boolean} relationships. In most of the approaches, a maximum number \iapcomment{$k$} of nodes which participate in the \iapcomment{boolean} functions of the network is assumed \textit{a priori} known, while for the learning to be computationally manageable by the method $k$ has to be very small, usually 4 or 5 even for middle-sized networks (\iapcomment{i.e 100 nodes}). Additional restrictions can pertain to the complexity of the functions, i.e, the nodes can interact only through AND or OR
operations.
\item Approaches which attempt to deal with both the parametric and logic part of the PBN, fail to model in a "compressed" way the transition probability matrix of the PBN and rely on the estimation of a large number of parameters. Therefore, they have unrealistic sample and runtime requirements.  
\item All of the methods do not report prediction accuracy for large (roughly more than 10 node) networks.
\end{enumerate}
The SCNFN is a compact (in terms of the involved parameters) representation of the PBN which enables an efficient, in terms of both sample and runtime demands, learning algorithm. The learning procedure circumvents all of the above limitations by greedily identifying the node interactions and approximately maximizing the likelihood of the observed temporal sequences. Subsequently, the SCNFN turns to be an efficient modeling structure in the sense that it can be efficiently sampled/simulated in order to obtain long-run transition probabilities of the generated Markov Chain. 
%
%
%
\appendix[Disjunction Learning]
Each recursive call of Algorithm $\ref{algo:scnf_disjunction_learn}$ inserts a new literal $l^{*}$ in the currently formed disjunction $\phi$ (see Lines 25 and 30). The inclusion of a new literal $l$ results in turning to $True$ some transitions in $H_T$ (this is desirable, since it reduces the number of recursive calls required for the new disjunction to satisfy the condition i in Line 7 which is guaranteed by  Line 24) and in $H_F$ (this is not desirable, since it increases the probability that the condition ii in Line 7 which is guaranteed by Line 21 is not satisfied, as well as the number of disjunctions that have to be included in the SCNF clause until the condition in Line 15, Algorithm \ref{algo:scnf_logic_learn} is satisfied). Therefore, for each candidate literal we compute a positive, see Line 13 (negative, see Line 14) score according to the number of positive (negative) transitions that will be satisfied (invalidated) if the literal is included in the disjunction, and a normalized total score, see Line 15. The literal with the maximum normalized score is selected (Line 17). \iapcomment{In Lines 18-20, it is scrutinized whether all available literals have non-positive score (no transition in $H_T$ will be removed after its inclusion). Note that it suffices to check only the positive score of the literal with the largest normalized score to verify this condition. If this is the case, the recursion has to stop since the current combination of literals in $\phi$ will result in an invalid disjunction (some transitions in $H_T$ may not be satisfied)}. Similarly, the current branch of the recursion leads to an invalid disjunction, if after the inclusion of the new literal all the transitions in $H_F$ are evaluated as $True$ by the $\phi \lor l^{*}$. If this is the case, the algorithm should continue with finding the next best literal (Line 22) by excluding $l^{*}$ from the set of the available literals which can be used for the formation of the remaining disjunction.  In case the inclusion of $l$ still preserves the validity of the disjunction $\phi \lor l^*$, $H_T$($H_F$) are \iapcomment{reduced} in Line 28 (Line 29) and the recursion proceeds by removing both $l^*$ and $\neg{l^*}$ from the set of the available literals.  If the condition in Line 31 is $False$, a valid disjunction is discovered and returned \iapcomment{(Line 33)}, otherwise the algorithm backtracks (Line 32) after $H_T$  and $H_F$ are restored and $l^{*}$ is removed from the set of the available literals that will be considered for inclusion in the next routine call. In the worst case, the algorithm will perform exhaustive search in the space of all possible disjunctions  until it discovers a valid formula. However, due to the heuristic score of literals as defined in Line 15, the number of backtracks becomes negligible and it does not affect the computational performance of the algorithm.
\allowdisplaybreaks
\begin{algorithm}[t]
\caption {$\textit{Disjunction-Learn}$}
\label{algo:scnf_disjunction_learn}
\begin{algorithmic}[1]
\scriptsize
\setstretch{0.8}
\State \textbf{Inputs} 
\State \ \ \ $H_F$: The set of negative transitions.
\State \ \ \ $H_T$: The set of positive transitions.
\State \ \ \ $L$: The set of the available literals.
\State \ \ \ $\phi$: currently formed disjunction.
\State \textbf{Output} 
\State \ \ \ {$\phi^{*}$: \iapcomment{A} new disjunction, which satisfies:

 i) $\phi^{*}(\mathbf{S})=True, \forall \mathbf{S} \in H_T$ 
 
 ii)$\exists \mathbf{S} \in H_F$}: $\phi(\mathbf{S})=False$
\State \textbf{Begin}  \newline

\State \textbf{If}($L== \emptyset$) 
\State \ \ \ \textbf{Return} $\emptyset$
\State  \textbf{EndIf} \newline

\State \textbf{For }($\forall l \in L$) 

\State \ \ \  $score^{+}(l)=|\{\mathbf{S} \in H_T: l(\mathbf{S})=True\}|$ \newline

\State \ \ \  $score^{-}(l)=|\{\mathbf{S} \in H_F: l(\mathbf{S})=True\}|$

\State \ \ \  $score(l)=\frac{score^{+}(l)}{|H_T|}-\frac{score^{-}(l)}{|H_F|}$

\State \textbf{EndFor} \newline

\State  $l^{*}\leftarrow \argmax_{l \in L} score(l)$ \newline

\State \textbf{If} $\big(score^{+}(l^{*})==0\big)$ 
\State \ \ \ \textbf{Return} $\emptyset$;
\State \textbf{EndIf}
\newline

\State  \textbf{If} $\big(score^{-}(l^{*})==|H_F|\big)$ 
\State \ \ \ \textbf{Return} $Disjunction-Learn\big(H_F,H_T,L-l^{*},\phi\big)$
\State \textbf{EndIf}
\newline

\State  \textbf{If } $\big(score^{+}(l^{*})==|H_T|\big)$ 
\State \ \ \ $\phi^{*}\leftarrow \phi \lor l^{*}$  
\State \ \ \ \textbf{Return} $\phi^{*}$
\State   \textbf{EndIf}
\newline

\State  $\hat{H}^{(T)}\leftarrow H_T- \left\{\mathbf{S}: \mathbf{S} \in H_T, l^{*}(\mathbf{S})=True\right\}$  
\newline

\State  $\hat{H}^{(F)}\leftarrow H_F- \left\{\mathbf{S}: \mathbf{S} \in H_F, l^{*}(\mathbf{S})=True\right\}$  
\newline

\State $\phi^{*} \leftarrow Disjunction-Learn\big(\hat{H}^{(F)}, \hat{H}^{(T)}, L-l^{*}-\neg{l^{*}}, \phi \lor l^{*}\big)$
\newline

\State \textbf{If}($\phi^{*}== \emptyset$) 
\State \ \ \ \textbf{Return} $Disjunction-Learn\big(H_F, H_T, L-l^{*}, \phi\big)$ \newline
\State \textbf{Return} $\phi^{*}$
\end{algorithmic}
\end{algorithm}
\section*{Acknowledgements}
The first author would like to acknowledge the "Alexander S. Onassis", Public Benefit Foundation,  for awarding a graduate fellowship . The first author also acknowledges support from ATK/Nick G. Vlahakis, Gerondellis Foundation and  the Carnegie Mellon University College of Engineering.
We also thank Han Zhao, Anthony Platanios, Filippe Condessa, Pengtao Xie and William Guss for the helpful discussions and reviews of this article.
\bibliographystyle{IEEEtran}
\bibliography{bare_jrnl}
\ifCLASSOPTIONcaptionsoff
  \newpage
\fi
\end{document}